\documentclass[10pt,journal,compsoc]{IEEEtran}

\ifCLASSOPTIONcompsoc
\else
\fi

\ifCLASSINFOpdf
\else
\fi

\usepackage{amsmath, amsthm, amssymb}
\usepackage{amssymb}
\usepackage{url,flushend,multirow,booktabs}
\usepackage{algorithm,algorithmic}
\usepackage{graphicx}
\usepackage{epstopdf}
\usepackage{epsfig}
\usepackage{bm}
\usepackage{subfigure}
\usepackage{float}
\usepackage{cite}

\newcommand{\tr}{\textrm{Tr}}
\newcommand{\eg}{e.g.}
\newcommand{\ie}{i.e.}
\newcommand{\wrt}{w.r.t.}
\newcommand{\etal}{et al.}

\renewcommand{\algorithmicrequire}{\textbf{Input:}}   
\renewcommand{\algorithmicensure}{\textbf{Output:}}  
\renewcommand{\bm}{\mathbf}
\DeclareFixedFont{\mf}{OT1}{ptm}{m}{n}{10pt}
\DeclareFixedFont{\mfb}{OT1}{ptm}{bx}{n}{10pt}

\begin{document}
%
\title{Constrained Low-rank Learning Using Least Squares Based Regularization}

\author{Ping~Li,~\IEEEmembership{Member,~IEEE,} 
        Jun~Yu,~\IEEEmembership{Member,~IEEE,}
        Meng~Wang,~\IEEEmembership{Member,~IEEE,}
        Luming~Zhang,~\IEEEmembership{Member,~IEEE,}
        Deng~Cai,~\IEEEmembership{Member,~IEEE,} 
        and~Xuelong~Li,~\IEEEmembership{Fellow,~IEEE,}
\IEEEcompsocitemizethanks{\IEEEcompsocthanksitem 
This work was supported in part by the National Natural Science Foundation of China under Grant 61502131, Zhejiang Provincial Natural Science Foundation of China under Grant LQ15F020012, the National Basic Research Program of China (973 Program) under Grant 2013CB336500, the National Natural Science Foundation of China under Grants 61572169, 61472266, 61472110, and China Scholarship Council.
\IEEEcompsocthanksitem Corresponding author: P.~Li (patriclouis.lee@gmail.com)}
\thanks{Manuscript received July 31, 2016.}
}

\markboth{For personal use only}
{LI \MakeLowercase{\textit{et al.}}:~CONSTRAINED LOW-RANK LEARNING USING LEAST SQUARES BASED REGULARIZATION}

\IEEEtitleabstractindextext{%
\begin{abstract}
Low-rank learning has attracted much attention recently due to its efficacy in a rich variety of real-world tasks, \eg, subspace segmentation, image categorization. Most low-rank methods are incapable of capturing low-dimensional subspace for supervised learning tasks, \eg, classification and regression. This work aims to learn both the discriminant low-rank representation and the robust projecting subspace in a supervised manner. To achieve this goal, we cast the problem into a constrained rank minimization framework by adopting the least squares regularization. Naturally, the data label structure tends to resemble that of the corresponding low-dimensional representation, which is derived from the robust subspace projection of clean data by low-rank learning. Moreover, the low-dimensional representation of original data can be paired with some informative structure by imposing an appropriate constraint, \eg, Laplacian regularizer. Therefore, we propose a novel \textit{Constrained Low-Rank Representation} (CLRR) method. The objective function is formulated as a constrained nuclear norm minimization problem, which can be solved by the inexact Augmented Lagrange Multiplier algorithm. Extensive experiments on image classification, human pose estimation and robust face recovery have confirmed the superiority of our method.
\end{abstract}

}

\maketitle

\IEEEdisplaynontitleabstractindextext

\IEEEpeerreviewmaketitle

\section{Introduction}
\label{sec1:intro}
Low-rank learning has exhibited its advantages in a broad range of real-world applications, such as subspace segmentation, image classification, outlier detection. Besides, it theoretically supports dealing with contaminated observations by outliers and noise, \eg, disguise, occlusion, specular reflections or pixel corruptions. Hence, it has been receiving much attention from both academia and industry, recently. As suggested in low-rank matrix recovery and matrix completion, we assume that data points from the same pattern tend to linearly correlate in the subspace. Thus, data points from different categories can be treated as samples nearly drawn from a union of multiple low-rank subspaces. However, the collected data points might be vulnerable to noise and corruptions in unfavorable situations, such as varying lighting, serious fading and partial occlusion, potentially damaging the subspace structures as well as deteriorating the learning process. Therefore, it is desirable to develop one technique to recover clean data from noisy observations while maintaining the intrinsic subspace structure of the data.

To this end, there have emerged a number of low-rank learning methods \cite{candes2011robustPCA,liu-pami2013-lrr,zhang-nn2014-similarity}. Among them, Robust Principal Component Analysis (RPCA) \cite{candes2011robustPCA} and Low-Rank Representation (LRR) \cite{liu-pami2013-lrr} are two typical approaches, which decouple the noisy data into the clean component (\ie, the recovered data) and the sparse error component. The difference is that RPCA implicitly assumes the underlying data structure is a single low-rank subspace. This neglects the specific property of the individual subspace. By contrast, LRR explicitly considers the multiple low-rank subspaces by adopting a dictionary linearly spanning the data space, thus better respecting the underlying data structure. In theory, LRR can be considered as a generalization of RPCA for which the dictionary degenerates to identity matrix. Due to their success in a vast number of applications, they are often refined to satisfy different requirements. For example, some researchers \cite{liang2012repairing,zhuang-cvpr2012-non} explored the idea of enforcing the sparsity constraint on the lowest rank representation while others \cite{liu-cvpr2012-fixed} adopted the fixed-rank strategy to accelerate the computation.

Nevertheless, these methods are unsupervised and cannot encode the prior knowledge (\eg, pairwise constraints, label structures) which could help capture the discriminant information. In this paper, we seek a robust projecting subspace, where the label structure of training data is inherently incorporated. It is expected that the label structure of the data could be well embedded in low-rank representation, which integrates the recovered data with discriminating power. Thus, we can not only use the recovered data to enjoy more promising classification performances, but also employ the projecting subspace to obtain better regression results. Furthermore, to equip the learned subspace with some appealing properties, we impose an adaptive constraint on the low-dimensional representation of original data as a regularizer. Several popular constraints can be used, \eg, the locality constraint \cite{pei-tcyb2014-automated} and the Fisher constraint \cite{huang-tnnls2012-semi}. Coupling these factors, we develop a novel method called \textit{Constrained Low-Rank Representation} (CLRR) to achieve the aforementioned goals. We formulate it as a constrained rank minimization problem, which can be solved by the inexact Augmented Lagrange Multiplier (ALM) algorithm \cite{lin-tr2009-alm}. The primary advantage of our method is that it can yield a robust projecting subspace where data points can be mapped into the low-dimensional data space, as well as generating the discriminant lowest rank representation of the data. To investigate the performance of our method, we have applied CLRR to several practical tasks, including image classification, human pose estimation, and robust face recovery. Empirical studies have shown its superiority over several alternatives, \eg, CLRR can achieve the lowest classification error rates on three publicly available image databases, and yields the lowest angle errors on 3D human body poses recovery.

The remainder of the paper is structured as follows. Firstly, Section~\ref{sec2:related} briefly reviews the previous works related to ours. Then, we describe the problem setting of this work in Section~\ref{sec3:problem} and introduce the proposed method covering the formulations, followed by its optimization framework in Section~\ref{sec4:method}. Moreover, we have some discussions in Section~\ref{sec5:discussion}. Furthermore, a number of comprehensive experiments were conducted on several real-world databases and the results are reported in Section~\ref{sec6:experiment}. Finally, we conclude the paper in Section~\ref{sec7:conclusion}.

\section{Related Work}
\label{sec2:related}
In this section, we mainly describe some works related to our proposed method, including low-rank learning, constrained matrix factorization and subspace learning.

Low-rank learning has attracted much attention due to its overwhelming advantages in a wide range of real-world applications \cite{chen-tcyb2014-robust, li-sigpro2015-sparse}, such as subspace segmentation, face recognition, outlier detection and video surveillance. Given a collection of data points, to consider the situation where a fraction of data matrix entries are missing or arbitrarily corrupted, Cand{\`e}s' \etal \cite{candes2011robustPCA} proposed Robust Principal Component Analysis (RPCA), which treats this problem as decomposing the input matrix into two components involving a low-rank matrix and a sparse matrix. However, it suffers from the deficit that data points are assumed to be populated in one subspace, which violates popular situations where data points reside on multiple subspaces. To overcome this shortcoming, Liu \etal \cite{liu-pami2013-lrr} put forward Low-Rank Representation (LRR) which was proved to be very effective for robust subspace segmentation. LRR seeks the low-rank component based on a dictionary, which could be learned from the data or simply be the data space. Hence, it is expected that LRR could better cater for the requirements of many real-world tasks, \eg, image restoration \cite{peng-tcyb2014-reweighted}. Actually, it shares with Sparse Subspace Clustering (SSC) \cite{elhamifar-pami2013-ssc} the common assumption that the underlying subspaces of the data are low-dimensional. The different philosophy lies in that SSC inherently seeks a sparse representation while LRR attempts to make the representation low-rank, leading to different objective functions. Besides, someone have tried to consider inducing both the sparsity and the low-rankness of the representation \cite{zhuang-cvpr2012-non}, so that the intrinsic structure of the underlying subspace could be well respected. Thereafter, Liu \etal \cite{liu-iccv2011-latent} proposed Latent LRR that exploits both the observed and the hidden data to construct the dictionary, in order to complement the insufficient observations. Moreover, to consider the intrinsically geometric structure of the data, Yin \etal \cite{yin-pami2016-laplacian} proposed a Laplacian regularized LRR model to learn nonnegative sparse and low-rank representation, which takes into account the high-order relations among data points by hypergraphs. Furthermore, to avoid the expensive singular value decomposition, Liu \etal \cite{liu-cvpr2012-fixed} adopted a fixed-rank strategy to obtain the low-rank representation, which largely improves the computational speed. Aside from that, one can extend the work to semi-supervised scenario by utilizing the prior knowledge in the form of constraints \cite{wang-tcyb2014-constraint}, and can employ LRR to substitute sparse code or vector quantization in spatial pyramid matching to encode local descriptors \cite{peng-kbs2015-fast}. In addition, some researchers use low-rank learning for dictionary learning, \eg, Jiang \etal \cite{jiang-pami2015-sparse} proposed supervised low-rank dictionary decomposition to facilitate a sparse and dense hybrid representation framework as well as alleviating the problem of the corrupted training data for face recognition; Li \etal \cite{li-ivc2014-learning} attempted to refine the sparse representation by learning a discriminative dictionary with one low-rank regularizer.

Generally speaking, low-rank representation can be cast into the paradigm of matrix factorization, which plays an important role in machine learning and pattern recognition. As is known to us, several constraints can be imposed onto matrix factorization to make the learned representation more informative. Therefore, constrained matrix factorization as a principled fashion has been widely applied in a large quantity of methods. Recently, one of the most popular constraints is graph-based regularization by virtue of manifold learning, which is capable of preserving the geometrical structure of the data space, such as graph-based non-negative matrix factorization methods \cite{pei-tcyb2014-automated} and concept factorization \cite{liu-tcyb2014-constrained}. Moreover, to better encode the manifold structure of the data, some researchers attempt to use multiple graphs to boost the learning performances, such as classification \cite{wu-tcyb2015-boosting,pan-tcyb2015-graph} and coclustering \cite{li-tcyb2013-relational}. And some others consider the problem from the topographic perspective, \eg, topographic non-negative matrix factorization for data representation \cite{xiao-tcyb2014-topographic}. Motivated by previous works, we have designed a regularizer that incorporates the informative constraints, such as local consistency and discriminative property, to enforce the effective structure onto the learned low-rank representation.

Actually, our proposed method can be also treated as a subspace learning method, since it aims to learn both the low-rank subspace and the robust projecting subspace. To this end, subspace learning has established itself as an important tool to obtain effective data representation in the last decade. Typical subspace learning methods have been proved to be very effective in practical applications, such as Principal Component Analysis (PCA) and Linear Discriminant Analysis (LDA) \cite{bishop-book2006-prml}. Recently, Jiang \etal \cite{jiang-pami2008-eigenfeature} developed a subspace method for facial eigenfeature regularization and extraction (ERE), and the eigenspace of the within-class scatter matrix is decomposed into three subspaces involving a reliable subspace spanned mainly by variation, an unstable one due to noise as well as limited training data, and a null subspace. This could alleviate the problem of instability, overfitting and poor generalization. As we know, PCA is an unsupervised method and does not require label information. To take into account the label information, an Asymmetric Principal Component Analysis (APCA) \cite{jiang-pami2009-asymmetric} approach was proposed, which utilizes class covariance matrices and enables removing the unreliable dimensions of principal components. While APCA is designed for handling the two-class problem, Supervised Principal Component Analysis (SPCA) \cite{jiang-spm2011-linear} deals with the multiple-class problem. Unlike APCA, SPCA imposes different weights on the covariance matrices so as to consider class-specific information of the data set. To summarize, the above methods are expected to generate considerable subspaces, but they cannot explicitly yield low-rank subspaces and separate the error matrix as our approach can, \eg, they cannot recover the clean component of the corrupted image. In some sense, this would hinder potentially more widespread applications of these methods.

It is worth noting that there exist two works related to ours. One is the Supervised Regularization based Robust Subspace (SRRS) method \cite{li-sdm2014-robust}, which smoothly integrates subspace learning and data recovery in a unified framework to jointly learn discriminative subspace and low-rank representation from the data. It differs from our method in several aspects: (a) it adopts the Fisher criterion to capture the discriminant structure while ours utilizes both the Laplacian regularizer and the least squares regularizer under the guidance of the supervised information; (b) it includes a generalized eigen-decomposition problem to obtain the projecting subspace while ours gives a closed-form solution accordingly and avoids solving the expensive Sylvester equation; (c) Our method can be used for regression tasks directly while SRRS cannot. The other is Robust Regression (RR) \cite{Huang-eccv2012-robust}, which leverages the rank regularizer and the sparse error term, but it regards the underlying data structure as a single low-rank subspace that might cause the inaccurate recovery. By contrast, we assume the data populates on a mixture of multiple subspaces to guarantee the correct recovery. Moreover, the subspace obtained from RR does not have the desired informative properties, \eg, the locality-preserving ability, while our method can easily achieve this based on the adaptive regularizer. Details of our method are elaborated in the following sections.

%
\section{Problem Setting}
\label{sec3:problem}
In this paper, we define the constrained low-rank learning problem as follows. Given a collection of data points $\{\mathbf{x}_1, \mathbf{x}_2, \ldots, \mathbf{x}_n\}$ and their labels $\{\text{y}_1, \text{y}_2, \ldots, \text{y}_n\}$ distributed in $k$ classes, we assume they are samples approximately drawn from a mixture of several subspaces \cite{liu-pami2013-lrr}. The principal goal is to seek the discriminant lowest rank representation $\mathbf{Z}$ as well as the robust projecting subspace $\mathbf{P}$. More specifically, we denote the training data by $\mathbf{X}\in \mathbb{R}^{d \times n}$ with each data point stacked in a column, and the data matrix can be decomposed into a clean component $\tilde{\mathbf{X}}=\mathbf{AZ}$ and an error component $\mathbf{E}\in \mathbf{R}^{d \times n}$, where $\mathbf{A}\in \mathbb{R}^{d \times m}$ is treated as the dictionary linearly spanning the data space while $\mathbf{Z} \in \mathbb{R}^{m \times n}$ reveals the underlying subspace structure of the data.

More importantly, we argue that the recovered data can be mapped onto a low-dimensional data space by the robust projecting subspace $\mathbf{P}\in \mathbb{R}^{d \times k}$ (the reduced dimension $k$ is set to the number of classes), \ie, $\mathbf{V} = \mathbf{P}^T\mathbf{AZ} \in \mathbb{R}^{k\times n}$. On one hand, the low-dimensional data representation $\mathbf{V}$ is expected to be closely correlated to the label indicator matrix $\mathbf{Y}\in \mathbb{R}^{k\times n}$ while it acts as the estimated output given the input data. The matrix $\mathbf{Y}$ takes discrete values for classification and continuous values for regression, respectively, \eg, the entries in each column of $\mathbf{Y}$ are set to 1 if the sample belongs to the corresponding class. On the other hand, it is easy to endow the low-dimensional representation $\mathbf{P}^T\mathbf{X}$, derived from the original data space, with several appealing properties like the locality-preserving ability, by the constraint matrix $\mathbf{L} \in \mathbb{R}^{n \times n}$. Usually, this matrix should be semi-positive definite to make the imposed regularizer convex.

By tradition, the lowest rank representation is employed to construct an affinity matrix for subspace segmentation in unsupervised learning. Here, we mainly use it for recovering the clean data by $\mathbf{AZ}$, where $\mathbf{Z}$ plays a dominant role. Under such circumstance, both the recovered training data and testing data could show the robustness to noise or corruptions, and it also allows to discriminate the samples from different categories.

\section{Our Method}
\label{sec4:method}
This section concentrates on elaborating the proposed method, including the formulation, and the optimization framework as well as the algorithmic procedures.
\subsection{Formulation}
As mentioned earlier, our goal is to jointly seek the discriminant lowest-rank representation $\mathbf{Z}\in \mathbf{R}^{m\times n}$ and the robust projecting subspace $\mathbf{P}\in \mathbb{R}^{d\times k}$ in a supervised manner. Essentially, we have to minimize rank($\mathbf{Z}$), which is yet difficult to solve due to its discrete nature. As a common practice in low-rank methods \cite{liu-pami2013-lrr,zhuang-cvpr2012-non}, we use the nuclear norm as its convex surrogate. In this work, the dictionary $\mathbf{A}$ is set to $\mathbf{X}$. Hence, our objective function can be formulated as:
\begin{equation} \label{eq:clrr-2}
\begin{split}
\min_{\bm{Z}, \bm{E}, \bm{P}}&~\|\mathbf{Z}\|_* + \lambda \|\mathbf{E}\|_{2,1} + \alpha \tr(\mathbf{P}^T\mathbf{XLX}^T\mathbf{P}) + \beta\|\mathbf{V} - \mathbf{Y}\|_F^2,  \\
\text{s.~t.} & ~\mathbf{X}=\mathbf{XZ}+ \mathbf{E}, \mathbf{V} = \mathbf{P}^T\mathbf{XZ}, \mathbf{1}_n^T\mathbf{Z} = \mathbf{1}_n^T.
\end{split}
\end{equation}
where the nuclear norm $\| \cdot \|_*$ is the sum of singular values of a matrix, the group sparse norm $\| \cdot \|_{2,1}$ computes the sum of absolute values of $l_2$-norm on each column vector of a matrix, \eg, $\sum_j \|\mathbf{E}_j\|_2$ for $\mathbf{E}$, $\| \cdot \|_F$ denotes the Frobenius norm of a matrix, $\mathbf{1}$ is a column vector with all ones. The parameter $\alpha>0$ balances the contribution of the constraint to the objective, $\beta>0$ controls the fitting of the least squares term, and $\lambda>0$ governs the noise level. Note that the regularized error component can also be replaced by $l_1$-norm often used for sparse coding \cite{wang-imm2016-collaborative} if necessary.

In the objective function, the first term aims to minimize the rank of $\mathbf{Z}$ while the second term encourages the sparseness of the error matrix $\mathbf{E}$ for different groups. The former two terms are generally used by most low-rank learning methods, while the latter constraints are coherently taken into account for the first time. The third term is the enforced Laplacian regularizer, which can make the derived low-dimensional representation $\mathbf{P}^T\mathbf{X}$ be characteristic with the intrinsic property of the constraint matrix $\mathbf{L}$. When $\mathbf{L}$ is simply set to the identity matrix $\mathbf{I}$, this term reduces to its Frobenius norm. The fourth term is the least squares regularizer, which would make the recovered low-dimensional representation $\mathbf{V}$ have the similar structure as that of the label matrix $\mathbf{Y}$. Besides, we explicitly impose the normalization constraint on the columns of $\mathbf{Z}$ to ease the non-unique solution problem \cite{liu-iccv2011-latent}. It should be noted that we explicitly use the supervised information in the least squares regularizer to guide the learning process of the projecting subspace $\mathbf{P}$ and the lowest-rank representation $\mathbf{Z}$, and this strongly encourages the discriminative power for them. Moreover, while the error matrix $\mathbf{E}$ is used to encode the noise or corruptions, it is still impossible to eliminate all noise or corruptions of data points in practice. As a result, it is sensible to impose constraints on the low-dimensional representation of the original data space $\mathbf{X}$. Thus, the supervised information plays an overwhelming role on the low-dimensional representation for both the original data space and the recovered data space by the third and the fourth constraints, respectively. In addition, the two constraints are said to be inherently correlated in between, since during the optimization process the matrix $\mathbf{P}$ and the matrix $\mathbf{Z}$ are simultaneously updated by iteration. Details are shown below.

\subsection{Optimization}
In this part, we show how to optimize the objective function in (\ref{eq:clrr-2}) by the Augmented Lagrange Multipler (ALM) algorithm \cite{lin-tr2009-alm}, which has a variant named the Alternating Direction Method (ADMM) \cite{lin-nips2011-adm} widely used in solving low-rank based problems \cite{li-sdm2014-robust,ye-cvpr2012-robust,liu-iccv2011-latent}.

To make the objective function separable and solvable, we introduce the relaxation variable $\mathbf{J}$ to represent $\mathbf{Z}$ and substitute the constraint $\mathbf{V}=\mathbf{P}^T\mathbf{XZ}$ into (\ref{eq:clrr-2}), leading to the equivalent problem:
\begin{equation} \label{eq:clrr-3}
\begin{split}
\min_{\bm{Z}, \bm{E}, \bm{P},\bm{J}}&~\|\mathbf{J}\|_* + \lambda \|\mathbf{E}\|_{2,1} + \mathcal{F}(\mathbf{Z}, \mathbf{P}),  \\
\text{s.~t.} & ~\mathbf{X}=\mathbf{XZ}+ \mathbf{E}, \mathbf{1}_n^T\mathbf{Z} = \mathbf{1}_n^T, \mathbf{Z} = \mathbf{J}.
\end{split}
\end{equation}
where the constraint term is:
\begin{equation}
\mathcal{F}(\mathbf{Z}, \mathbf{P}) = \alpha\tr[\mathbf{P}^T\mathbf{XLX}^T\mathbf{P}] + \beta\|\mathbf{P}^T\mathbf{XZ} - \mathbf{Y}\|_F^2.
\end{equation}

The above problem in (\ref{eq:clrr-3}) can be solved by minimizing its augmented Lagrangian function:
\begin{equation} \label{eq:obj}
\begin{split}
\mathcal{L} = &~
\|\mathbf{J} \|_{*} + \lambda \|\mathbf{E}\|_{2,1} + \mathcal{F}(\mathbf{Z},\mathbf{P})\\
+ &~\tr[\mathbf{\Psi}_a^T(\mathbf{X}-\mathbf{XZ}- \mathbf{E})] \\
+ &~ \tr[\mathbf{\Psi}_b^T(\mathbf{Z}-\mathbf{J})] + \tr[\mathbf{\Psi}_c^T(\mathbf{1}_n^T\mathbf{Z}-\mathbf{1}_n^T)] \\
+ &~ \frac{\mu}{2}(\|\mathbf{X}-\mathbf{XZ}- \mathbf{E}\|_F^2 + \| \mathbf{Z}-\mathbf{J}\|_F^2 \\
+ &~\|\mathbf{1}_n^T\mathbf{Z}-\mathbf{1}_n^T\|_F^2 ),
\end{split}
\end{equation}
where the matrices $\mathbf{\Psi}_a\in \mathbb{R}^{d\times n}$, $\mathbf{\Psi}_b\in \mathbb{R}^{n\times n}$, $\mathbf{\Psi}_c\in \mathbb{R}^{1\times n}$ are the Lagrange multipliers, and $\mu>0$ is a penalty parameter.  Now, this problem becomes an unconstrained one, which can be solved using an alternating strategy. In other words, we can respectively minimize the variables $\mathbf{Z}$, $\mathbf{E}$, $\mathbf{P}$, $\mathbf{J}$ by holding the rest, followed by updating the Lagrange multipliers $\mathbf{\Psi}_a$, $\mathbf{\Psi}_b$, $\mathbf{\Psi}_c$.  The convergence of the inexact ALM algorithm has been provably guaranteed with mild conditions \cite{lin-tr2009-alm} when optimizing multiple variables.

\textbf{Calculation of $\mathbf{P}$:}
By fixing $\mathbf{Z}$, $\mathbf{E}$, $\mathbf{J}$ and dropping the constant terms, (\ref{eq:obj}) can be rewritten as:
\begin{equation} \label{eq:obj-P}
\mathcal{L}(\mathbf{P}) = \mathcal{F}(\mathbf{P}).
\end{equation}
Taking its derivative and let $\nabla\mathcal{L}(\mathbf{P}) = 0$, we can easily obtain the closed-form solution $\mathbf{P}$:
\begin{equation} \label{eq:obj-P2}
\mathbf{P} =[\mathbf{X}(\alpha\mathbf{L}+\beta\mathbf{ZZ}^T)\mathbf{X}^T]^{-1} \mathbf{XZY}^T.
\end{equation}
This step is more efficient than that in \cite{li-sdm2014-robust}, which needs to solve the expensive generalized eigen-decomposition problem.

\textbf{Calculation of $\mathbf{J}$:}
By holding the variables $\mathbf{P}$, $\mathbf{E}$, $\mathbf{Z}$ and eliminating the irrelevant terms, (\ref{eq:obj}) degenerates to:
\begin{equation} \label{eq:obj-J}
\begin{split}
\mathcal{L}(\mathbf{J})=&~\|\mathbf{J} \|_{*} + \tr[\mathbf{\Psi}_b^T(\mathbf{Z}-\mathbf{J})]+ \frac{\mu}{2} \| \mathbf{Z}-\mathbf{J}\|_F^2 \\
=&~\frac{1}{\mu}\|\mathbf{J} \|_{*} + \frac{1}{2}\|\mathbf{J}-(\mathbf{Z}+ \frac{1}{\mu}\mathbf{\Psi}_b)\|_F^2.
\end{split}
\end{equation}
We employ the Singular Value Thresholding (SVT) operator \cite{cai-siam2010-svt} to compute the optimal $\mathbf{J}$ efficiently. Specifically, we first conduct Singular Value Decomposition (SVD) on the matrix $\mathbf{Z}+ \frac{1}{\mu}\mathbf{\Psi}_b=\mathbf{U}_J\mathbf{\Sigma}_J\mathbf{V}_J$, where $\mathbf{\Sigma}_J$ is a diagonal matrix with its entries being a group of singular values $\{{\sigma_i}\}_{i=1}^r$ ($r$ is the rank). Thereafter, we can obtain the optimal solution $\mathbf{J}^*=\mathbf{U}_J\mathbf{\Omega}_{\frac{1}{\mu}}\mathbf{\Sigma}_J\mathbf{V}_J$, where $\mathbf{\Omega}_{\frac{1}{\mu}}\mathbf{\Sigma}_J=\text{diag}(\{\sigma_i- \frac{1}{\mu}\}_+)$, and the marker ``+" denotes the positive part.

\textbf{Calculation of $\mathbf{Z}$:}
While keeping the variables $\mathbf{P}$, $\mathbf{E}$, $\mathbf{J}$ fixed, (\ref{eq:obj}) reduces to:
\begin{equation} \label{eq:obj-Z}
\begin{split}
 \mathcal{L} (\mathbf{Z})
 =&~\mathcal{F}(\mathbf{Z},\mathbf{P}) \\
+ &~\tr[\mathbf{\Psi}_a^T(\mathbf{X} - \mathbf{XZ}) + \mathbf{\Psi}_b^T\mathbf{Z} + \mathbf{\Psi}_c^T\mathbf{1}_m^T\mathbf{Z}] \\
+ &~ \frac{\mu}{2}(\|\mathbf{X}-\mathbf{XZ}- \mathbf{E}\|_F^2 + \| \mathbf{Z}-\mathbf{J}\|_F^2 \\
+ &~ \|\mathbf{1}_m^T\mathbf{Z}-\mathbf{1}_n^T\|_F^2).
\end{split}
\end{equation}
Now let the derivative of (\ref{eq:obj-Z}) \wrt~$\mathbf{Z}$ be zero, we have
\begin{equation} \label{eq:obj-Z2}
\mathbf{U}_a\mathbf{Z}=\mathbf{U}_b+\mathbf{U}_c,
\end{equation}
where
\begin{equation}
\mathbf{U}_a = 2\beta\mathbf{X}^T\mathbf{PP}^T\mathbf{X}+\mu(\mathbf{X}^T\mathbf{X}+ \mathbf{1}_n\mathbf{1}_n^T + \mathbf{I}_n),
\end{equation}
\begin{equation}
\mathbf{U}_b = 2\beta\mathbf{X}^T\mathbf{PY}+
\mu(\mathbf{X}^T\mathbf{X}-\mathbf{X}^T\mathbf{E}+\mathbf{J}+\mathbf{1}_n\mathbf{1}_n^T),
\end{equation}
\begin{equation}
\mathbf{U}_c = \mathbf{X}^T\mathbf{\Psi}_a-\mathbf{\Psi}_b-\mathbf{1}_n\mathbf{\Psi}_c.
\end{equation}
Then, we can obtain the solution to $\mathbf{Z}$, \ie,
\begin{equation} \label{eq:obj-Z3}
\mathbf{Z} = \mathbf{U}_a^{-1}(\mathbf{U}_b + \mathbf{U}_c).
\end{equation}
where the variable $\mathbf{U}_a$ is positive definite, which would make its inversion more stable to some extent during the solving process.

\textbf{Calculation of $\mathbf{E}$:}
Similarly to the above routines, we fix the variables $\mathbf{P}$, $\mathbf{J}$, $\mathbf{Z}$ and drop the constant matrices,  then (\ref{eq:obj}) can be reformulated as
\begin{equation} \label{eq:obj-E}
\begin{split}
\min_{\mathbf{E}} \frac{\lambda}{\mu}\|\mathbf{E}\|_{2,1} + \frac{1}{2}\|\mathbf{E} - (\mathbf{X}-\mathbf{XZ}+\frac{1}{\mu}\mathbf{\Psi}_a)\|_F^2.
\end{split}
\end{equation}
This problem has been solved by existing work \cite{liu-pami2013-lrr} and its optimal solution is given by
\begin{equation}  \label{eq:obj-E2}
\mathbf{E}(:, i) = \left\{\begin{array}{ll}
\frac{\|\Psi_i\|-\lambda}{\|\Psi_i\|}\Psi_i, & \text{$\frac{\lambda}{\mu}<\|\Psi_i\|$} ,   \\
0,&  \text{otherwise} .
\end{array}\right.
\end{equation}
where $\Psi_i$ is the $i$-th column vector of the matrix $\mathbf{\Psi}_a$. Up to now, the solutions to all four variables have been obtained in the optimization framework, and the complete procedures are summarized in Algorithm~\ref{alg1}. This framework can be used for both classification and regression tasks.
%
%
\begin{algorithm}[t]
 \caption{Solving Problem (\ref{eq:obj}) by Inexact ALM}
 \label{alg1}
 \begin{algorithmic}[1]
  \REQUIRE
  The data matrix $\mathbf{X}$, the label matrix $\mathbf{Y}$, the parameters $\alpha, \beta, \lambda$.

  \INITIALIZE
  Set all entries of $\mathbf{Z}$, $\mathbf{J}$, $\mathbf{E}$, $\mathbf{\Psi}_a$, $\mathbf{\Psi}_b$, $\mathbf{\Psi}_c$ to zero, $\mu=0.01$, $\mu_{max}=10^{6}$, $\rho=1.3$, $\epsilon=10^{-7}$.

  \PROCEDURE~~\\
  \WHILE{not converged}
  \STATE Fix the others and update $\mathbf{P}$, $\mathbf{J}$, $\mathbf{Z}$ and  $\mathbf{E}$, respectively, by solving (\ref{eq:obj-P2}), (\ref{eq:obj-J}), (\ref{eq:obj-Z}) and (\ref{eq:obj-E2}).
  \STATE Update the Lagrange multipliers:
  \begin{equation*}
  \begin{split}
  \mathbf{\Psi}_a &\leftarrow \mathbf{\Psi}_a + \mu (\mathbf{X}-\mathbf{XZ} - \mathbf{E}), \\
  \mathbf{\Psi}_b &\leftarrow \mathbf{\Psi}_b + \mu (\mathbf{Z}-\mathbf{J} ),\\
  \mathbf{\Psi}_c &\leftarrow \mathbf{\Psi}_c + \mu (\mathbf{1}_m^T\mathbf{Z}-\mathbf{1}_n^T).
   \end{split}
  \end{equation*}
  \STATE Update the parameter $\mu$: $\mu  \leftarrow \min(\rho\mu, \mu_{max})$.
  \STATE Check the convergence conditions: \\
  \vspace{-6pt}
  \begin{equation*}
  \begin{split}
  \|\mathbf{X}-\mathbf{XZ} - \mathbf{E}\|_{\infty} < \epsilon, \\
  \|\mathbf{Z}-\mathbf{J}\|_{\infty} < \epsilon,
  \|\mathbf{1}_m^T\mathbf{Z}-\mathbf{1}_n^T\|_{\infty} < \epsilon.
    \end{split}
  \end{equation*}
  \ENDWHILE
 \ENSURE
  $\mathbf{Z}, \mathbf{E}, \mathbf{P}$.\\
 \end{algorithmic}
\end{algorithm}  \vspace{-2mm}
%
\begin{algorithm}[!t]
 \caption{CLRR for Data Classification}
 \label{alg-classification}
 \begin{algorithmic}[1]
  \REQUIRE 
  The training data $\mathbf{X}$ and their labels $\mathbf{Y}$, the testing data $\mathbf{X}_t$.
  \STATE Utilize Algorithm~\ref{alg1} to derive the low-rank matrix $\mathbf{Z}$ for training data.
  \STATE Learn the low-rank matrix $\mathbf{Z}_t$ for testing data using degenerated Algorithm~\ref{alg1} while setting $\alpha=0$ and $\beta=0$.
  \STATE Recover the clean data by $\mathbf{XZ}$ and $\mathbf{X}_t\mathbf{Z}_t$ from corrupted training and testing data in respective.
  \STATE Train a model using a classifier and predict the classes of the testing data points.
  \end{algorithmic}
\end{algorithm}
%
%
\section{Discussion}
\label{sec5:discussion}
In this section, we concentrate on discussing the advantages, the constraints, and the computational complexity of the proposed CLRR method.

\subsection{Advantages}
As we know, CLRR explicitly takes advantage of supervised information, \ie, data labels, to guide the low-rank learning and the robust subspace projection. This property naturally differentiates it from previous works \cite{candes2011robustPCA,cai-kdd2013-equivalent,liu-pami2013-lrr,li-sdm2014-robust}. In this paper, we apply our method into several important real-world applications, \ie, data classification, pose estimation, and robust data recovery.
For classification in Algorithm \ref{alg-classification}, we can either use the recovered data $\mathbf{XZ}$ or the reduced data representation $\mathbf{P}^T\mathbf{XZ}$. However, in practice we found the classification performance of using $\mathbf{P}^T\mathbf{XZ}$ is less satisfying than that of using $\mathbf{XZ}$, especially when the number of classes is small. This could be attributed to the fact that the dimension of the former is fixed by the number of classes in the training data, which may result in some information loss to low-rank based representation, while the latter does not. For regression tasks, \eg, pose estimation in Algorithm \ref{alg-pose}, we use the robust projecting subspace $\mathbf{P}$ to estimate the outputs of testing data points, as shown in Algorithm \ref{alg-recovery}. For robust data recovery, we directly employ $\mathbf{XZ}$ to recover the clean component from corrupted data.

Since we enforce the low-rank representation and the robust projecting subspace to be intrinsically correlated with the data ground-truth by constraints, it will help yield discriminant representations, encouraging improved classification and regression performances. Meanwhile, the robust recovery is guided by the least squares constraint, allowing to better identify the noise in different classes. However, there exits a drawback in our method, \ie, it cannot directly recover the clean component of the testing data due to the lack of labels. Yet, to handle this, we can solve the degenerated objective function by simply dropping the third and the fourth regularization terms in (\ref{eq:clrr-2}).
%
\begin{algorithm}[!t]
 \caption{CLRR for Human Pose Estimation}
 \label{alg-pose}
 \begin{algorithmic}[1]
  \REQUIRE 
  The training motion captured data $\mathbf{X}$, the test data $\mathbf{X}_t$, the matrix $\mathbf{Y}$ spanned by pose vectors (\eg, body joint angles).
   \STATE Learn the projecting subspace $\mathbf{P}$ using Algorithm~\ref{alg1}.
  \STATE Estimate the pose of the test data by calculating $\mathbf{P}^T\mathbf{X}_t$, where each test sample is stacked in the column of $\mathbf{X}_t$.
 \end{algorithmic}
\end{algorithm}
%
\begin{algorithm}[!t]
 \caption{CLRR for Robust Data Recovery}
 \label{alg-recovery}
 \begin{algorithmic}[1]
  \REQUIRE 
  The contaminated data $\mathbf{X}$, the label matrix $\mathbf{Y}$.
   \STATE Learn the matrices $\mathbf{Z}$ and $\mathbf{E}$ through Algorithm~\ref{alg1}.
  \STATE Recover the clean data by computing $\tilde{\mathbf{X}}=\mathbf{XZ}$.
  \STATE Eliminate the noise existing in the contaminated data by the matrix $\mathbf{E}$.
 \end{algorithmic}
\end{algorithm}

\subsection{Constraints}
In all, there are two constraints in the objective function, \ie, the least squares regularizer and the Laplacian regularizer. Recall that the least squares constraint in our method allows to build the inter-connections among the discriminant lowest rank representation, the robust projecting subspace and the label structure of the training data. This actually encourages the supervised guidance when recovering the clean samples and mapping the data points onto the robust low-dimensional subspace. Moreover, the adaptive constraint on the low-dimensional representation can be equipped with the locality-preserving ability by manifold learning or the discriminating ability using the Fisher rule \cite{belhumeur-pami1997-eigenfaces}. On the whole, these properties can be unified into a regularizer from a graph viewpoint.

For instance, we consider a general regularizer in terms of the trace ratio criterion \cite{huang-tnnls2012-semi}. Given $n$ data points with labels, each point is treated as a vertex and the relation between two vertices are encoded by an edge weight. Define two similarity matrices including the within-class matrix $\mathbf{W}_w\in \mathbb{R}^{n\times n}$ and the between-class matrix $\mathbf{W}_b\in\mathbb{R}^{n\times n}$, in addition to the adjacency matrix  $\mathbf{W}_n\in\mathbb{R}^{n\times n}$, where $W_w(i, j) = \frac{1}{n_c}$ ($n_c$ is the number of samples in the $c$-th class) and $W_b(i, j)=\frac{1}{n}-\frac{1}{n_c}$ if $\mathbf{x}_i$ and $\mathbf{x}_j$ belong to class $c$, otherwise $W_w(i, j) = 0$ and $W_b(i, j)=\frac{1}{n}$; $W_n(i, j) =1$ if $\mathbf{x}_i$ and $\mathbf{x}_j$ are nearest neighbours, otherwise zero. Thus, their corresponding Laplacian matrices are $\mathbf{L}_w=\mathbf{D}_w-\mathbf{W}_w$, $\mathbf{L}_b=\mathbf{D}_b-\mathbf{W}_b$ and $\mathbf{L}_n=\mathbf{D}_n-\mathbf{W}_n$,  where the entries of the diagonal matrices $\mathbf{D}$ are the column or row sums of the weight matrices $\mathbf{W}$.  As shown in \cite{nie-aaai2008-trace}, the sum of the within-class scatter matrix $\mathbf{S}_w=\mathbf{P}^T\mathbf{XL}_w\mathbf{X}^T\mathbf{P}$ and the between-class scatter matrix $\mathbf{S}_b=\mathbf{P}^T\mathbf{XL}_b\mathbf{X}^T\mathbf{P}$ is the total-scatter matrix $\mathbf{S}_t = \mathbf{S}_w+ \mathbf{S}_b$ \cite{bishop-book2006-prml}, leading to $\mathbf{L}_t = \mathbf{L}_w + \mathbf{L}_b=\mathbf{I}_n-\frac{1}{n}\mathbf{11}^T$, which is a centering matrix. These Laplacian matrices are all positive semi-definite (PSD), and thus the sum of them is also PSD. They can be easily incorporated into our adaptive regularizer as an alternative of $\mathbf{L}$, which could be also a graph-based matrix \cite{pei-tcyb2014-automated}. In empirical studies, we adopted the between-class scatter matrix $\mathbf{S}_b$ as the Laplacian regularizer. Additionally, CLRR can be extended to the semi-supervised scenario by modifying the adaptive regularizer as in \cite{cai-iccv2007-sda,huang-tnnls2012-semi} where the prior knowledge is used as the supervised information. Apart from that, to better encode the low-rank representation, we can resort to a more informative dictionary learned from the data points iteratively as in \cite{zhang-cvpr2013-learning}.

\subsection{Computational Complexity Analysis}
For computational cost, we use the big $\mathcal{O}$ notation to express the computational complexity of the proposed method. In total, the objective function has four variables, whose solutions require iterative computing in the whole process. To update $\mathbf{P}$, it needs $\mathcal{O}(n^3)$ to compute $\mathbf{ZZ}^T$, which makes it be the most costing component in solving $\mathbf{P}$. To update $\mathbf{J}$, it needs $\mathcal{O}(n^2r+r^2n)$, where $r$ is the rank of $(\mathbf{Z}+ \frac{1}{\mu}\mathbf{\Psi}_b)$. To update $\mathbf{Z}$, it needs $\mathcal{O}(n^2d+ndk)$ to compute $\mathbf{U}_a$, $\mathbf{U}_b$ and $\mathcal{O}(n^2d)$ to compute $\mathbf{U}_c$. To update $\mathbf{E}$, it needs $\mathcal{O}(d^2)$ to compute $\mathbf{E}$. Suppose the updates stop after $t$ iterations, the overall cost for CLRR is
\begin{equation} \label{eq:cost}
\mathcal{O}[t(n^3+n^2(r+d)+r^2n+d^2+ndk)].
\end{equation}

In practical tests, our objective function usually converges fast, and thus $t$ is actually small. If the given data matrix is highly low-rank, \eg, many samples are linearly correlated, then we have $r\ll min(n,d)$. Besides, the number of classes $k$ can be omitted compared to $n$. Therefore, (\ref{eq:cost}) can be compacted into $\mathcal{O}(n^3+n^2d+d^2+nd)$. Hence, the computational cost of CLRR is at the same level of SRRS \cite{li-sdm2014-robust} which also requires to compute $\mathbf{ZZ}^T$. Moreover, during the iterations, SRRS requires to solve the Sylvester equation \cite{sorensen2002sylvester}, which is very expensive and the solution would be unstable occasionally. The cost could be further reduced when the number of data points is much less than that of features, \ie, $n\ll d$. While both of them are more computationally intensive than several alternatives, \eg, LRR and SPCA, they are equipped with more inspiring advantages, such as least squares guidance, Fisher information and locality-preserving property, which could strengthen the informative structures of the recovered data space and the robust projecting subspace.

\section{Experiments}
\label{sec6:experiment}
In this section, we have conducted a broad range of experiments to examine the performance of the proposed CLRR method in three real-world applications, \ie, image classification, human pose estimation and robust face recovery. All experiments were carried out in MATLAB R2014a on Windows 7 with Intel i7-5820K CPU at 3.30GHz.
%
\begin{table*}[!t]
\centering
\caption{Classification error rates (Dimension) on testing data using GIST feature.}
\label{tbl-res-gist}
\begin{tabular}{|l|c c c|c|l|c c c|}
\hline 
\multirow{2}{*}{Method} &
\multicolumn{3}{c|}{LibSVM + GIST (512D)} &
\multicolumn{1}{c|}{\multirow{2}{*}{}}&
\multicolumn{1}{c|}{\multirow{2}{*}{Method}}&
\multicolumn{3}{c|}{MMD + GIST (512D)} \\
\cline{2-4} \cline{7-9}
  & COIL100 & VOC2012 & Caltech101 & \multicolumn{1}{c|}{} & \multicolumn{1}{c|}{} & COIL100 & VOC2012 & Caltech101 \\
\hline \hline
ERE    & 0.0086 ~(500)    & 0.5362 ~(500)  & 0.3290 ~(500)  &   & ERE   & 0.0248 ~(500)  & 0.6746 ~(450)  & 0.3681 ~(512)   \\
SPCA   & 0.0081 ~(350)    & 0.5348 ~(300)  & 0.3262 ~(200)  &   & SPCA  & 0.0243 ~(500)  & 0.6763 ~(512)  & 0.3453 ~(150)   \\
RPCA   & 0.0081 ~(500)    & 0.5245 ~(500)  & 0.3257 ~(500)  &   & RPCA  & 0.0233 ~(512)  & 0.6517 ~(500)  & 0.3543 ~(512)   \\
LRR    & 0.0086 ~(500)    & 0.5258 ~(500)  & 0.3125 ~(500)  &   & LRR   & 0.0268 ~(150)  & 0.6488 ~(300)  & 0.3857 ~(150)   \\
FRR    & 0.0100 ~(500)    & 0.5368 ~(500)  & 0.3240 ~(500)  &   & FRR   & 0.0233 ~(200)  & 0.6506 ~(300)  & 0.3865 ~(200)   \\
LLRR   & 0.0133 ~(500)    & 0.5141 ~(500)  & 0.3017 ~(500)  &   & LLRR  & 0.0371 ~(350)  & 0.6469 ~(450)  & 0.3643 ~(450)   \\
SRRS   & 0.0062 ~(500)    & 0.5097 ~(500)  & 0.3081 ~(500)  &   & SRRS  & 0.0219 ~(200)  & 0.6432 ~(400)  & 0.3559 ~(400)   \\
CLRR  &\bf{0.0041} ~(512) & \bf{0.4912} ~(350) &\bf{0.2816} ~(500) & & CLRR & \bf{0.0185} ~(150)  & \bf{0.5936} ~(200)  & \bf{0.3235} ~(100)   \\
\hline 
\end{tabular}
\end{table*}
%
\begin{table*}[!t]
\centering
\caption{Classification error rates (Dimension) on testing data using LBP feature.}
\label{tbl-res-lbp}
\begin{tabular}{|l|c c c|c|l|c c c|}
\hline 
\multirow{2}{*}{Method} &
\multicolumn{3}{c|}{LibSVM + LBP (1239D)} &
\multicolumn{1}{c|}{\multirow{2}{*}{}}&
\multicolumn{1}{c|}{\multirow{2}{*}{Method}}&
\multicolumn{3}{c|}{MMD + LBP (1239D)} \\
\cline{2-4} \cline{7-9}
  & COIL100 & VOC2012 & Caltech101 & \multicolumn{1}{c|}{} & \multicolumn{1}{c|}{} & COIL100 & VOC2012 & Caltech101 \\
\hline \hline
ERE    & 0.0074 (1100)  & 0.5492 (1239)  & 0.3953 (1200)  &   & ERE   & 0.0067 (~850) & 0.6123 (1239)  & 0.4177 (1000)   \\
SPCA   & 0.0079 (~250)  & 0.5490 (~700)  & 0.3750 (~650)  &   & SPCA  & 0.0067 (~650) & 0.6123 (1239)  & 0.3870 (~650)   \\
RPCA   & 0.0084 (1150)  & 0.5296 (1239)  & 0.3717 (1200)  &   & RPCA  & 0.0043 (~700) & 0.6279 (1150)  & 0.4240 (1050)   \\
LRR    & 0.0084 (1200)  & 0.5360 (1239)  & 0.3607 (1200)  &   & LRR   & 0.0129 (~150) & 0.5931 (~150)  & 0.4421 (~150)   \\
FRR    & 0.0079 (1150)  & 0.5460 (1239)  & 0.3665 (1150)  &   & FRR   & 0.0133 (~200) & 0.5900 (~150)  & 0.4118 (~150)   \\
LLRR   & 0.0084 (1000)  & 0.5292 (1239)  & 0.3541 (1150)  &   & LLRR  & 0.0076 (~350) & 0.5648 (~550)  & 0.3909 (~450)   \\
SRRS   & 0.0079 (1000)  & 0.5334 (~800)  & 0.3688 (1200)  &   & SRRS  & 0.0071 (~200) & 0.5705 (~550)  & 0.4059 (~450)   \\
CLRR   & \bf{0.0067} (1000)  & \bf{0.5007} (~250)  & \bf{0.3354} (1239)  &   & CLRR  & \bf{0.0052} (~~50) & \bf{0.5463} (~550)  & \bf{0.3723} (~300)   \\
\hline 
\end{tabular}
\end{table*}

\subsection{Image Classification}
In this part, we investigate classification performances of several state-of-the-art methods on three publicly available image databases, \ie, COIL100, VOC2012, Caltech101.

\textbf{Database descriptions:}~\textbf{COIL100}\footnote{http://www.cs.columbia.edu/CAVE/software/softlib/coil100.php} contains 7,200 samples and 100 categories, each of which has 72 samples; \textbf{VOC2012}\footnote{http://pascallin.ecs.soton.ac.uk/challenges/VOC/voc2012/index.html} comes from the Visual Object Classes challenge 2012 \cite{everingham-ijcv2010-pascal}, which has 17,125 samples covering 20 different categories; \textbf{Caltech101}\footnote{http://www.vision.caltech.edu/Image\_Datasets/Caltech101/} consists of 8,677 pictures from 101 categories. Here, the background and clutter classes are abandoned. To represent the images, we adopt the toolbox used in \cite{khosla-iccv2013-modifying} to extract their feature descriptors, \ie, GIST, LBP, dense HOG, dense SIFT. GIST \cite{oliva-ijcv2001-gist} describes the spatial envelope of the image and has 512D; LBP \cite{ojala-pami2002-lbp} extracts the non-uniform local binary pattern and concatenates 3 levels of spatial pyramid to obtain final 1239D feature vector; dense HOG extracts HOG \cite{dalal-cvpr2005-hog} in a dense manner and concatenates $2\times2$ cells to obtain a descriptor at each grid; dense SIFT extracts SIFT \cite{lowe-ijcv2004-sift} in a dense manner at multiple patch sizes. For HOG and SIFT, we apply the bag-of-words and spatial pyramid pipeline to obtain the final feature vector, \ie, the LLC \cite{wang-cvpr2010-llc} framework using three layers with max pooling. To train the dictionary, we randomly select ten percent of the images or at least 20 images in each category to detect interest points, from which we used one million descriptors for clustering. The dictionary sizes of HOG and SIFT are 256 and 1,024, while their feature dimensions are 5,376 and 21,504, respectively.

\textbf{Performance comparisons:} We have investigated classification performances of the proposed CLRR method, ERE \cite{jiang-pami2008-eigenfeature}, SPCA \cite{jiang-spm2011-linear}, RPCA \cite{candes2011robustPCA}, LRR \cite{liu-pami2013-lrr}, FRR \cite{liu-cvpr2012-fixed}, Latent LRR (LLRR) \cite{liu-iccv2011-latent}, and SRRS \cite{li-sdm2014-robust}. Among them, ERE and SPCA are not low-rank subspace methods, but they can be both used to enhance classification performance by dimensionality reduction as CLRR does. Since APCA \cite{jiang-pami2009-asymmetric} is a degenerated version of SPCA while ERE has been shown to perform better than Fisher LDA \cite{jiang-pami2008-eigenfeature}, we thus compare CLRR with ERE and SPCA. The rest are all low-rank learning methods, being ideal alternatives to the proposed approach.

\textbf{Experimental setups:} Each database was randomly divided into three disjoint parts: training data ($50\%$), validation data ($20\%$), and testing data ($30\%$). For all methods, best parameters were empirically searched by two-fold cross validation on the validation data. For ERE, the start point of the noise region was estimated by varying $\mu$ from 0 to 4 with an interval of 0.5. For SPCA, there are two weights $\alpha_i$ and $\eta$ in the covariance mixture, and $\alpha_i$ was inversely proportional to the sample size of class $i$ \cite{jiang-pami2009-asymmetric} while the optimal $\eta$ was chosen from $2^{[-5:1:5]}$ to differentiate the covariance matrix $\mathbf{\Sigma}_{\alpha}$ from the total scatter matrix with $\eta=1$ as in \cite{jiang-spm2011-linear}. Suppose $d$ was the feature dimension, we selected the optimal $\lambda$ from $[0.2:0.2:2]/\sqrt{d}$ for RPCA, $[1:3:30]/\sqrt{d}$ for LRR and FRR, respectively. There are two parameters in LLRR and SRRS. For LLRR, we chose the optimal $\alpha$, $\lambda$ from $[0.6:0.2:1.4]$. For SRRS, we chose the optimal $\alpha$, $\lambda$ from $2^{[-4:2:4]}$. There are three parameters in CLRR. We chose the best $\alpha$ from $2^{[-8:1:3]}$, $\beta$ from $2^{[-3:1:5]}$, and $\lambda$ from $2^{[-5:1:3]}$. When there are more than one parameters, we adopted the one-by-one strategy, \eg, choose the best $\lambda$ for LLRR while fixing $\alpha$ to 1, and then choose the best $\alpha$ using the optimal $\lambda$. Parameter selections were done on the validation data. We did the best to choose optimal parameters for each method, and the parameter intervals vary for different methods. Actually, they could be further tuned to refine the results. In some sense, the compared methods can be treated as subspace approaches for image classification, and the optimal dimensionality of the learned data representation is difficult to know. CLRR is also considered as a subspace learning method in this scenario. In particular, the learned data representation $\mathbf{XZ}$ consists of one basis matrix $\mathbf{X}$ (dictionary) and one coefficient matrix $\mathbf{Z}$ (low-rank subspace). Though it can directly use the full low-rank subspace $\mathbf{Z}$ for training the classification model, we empirically found the performance usually reaches optimal using less dimensions rather than the whole. Hence, we have varied the dimension from 50 to the full with a grid of 50 for all methods when conducting classification. Note that it would be possible that multiple dimensions all yield the optimal results, and we report those with less dimensions.

\begin{table*}[!t]
\centering
\caption{Classification error rates (Dimension) on testing data using HOG feature.}
\label{tbl-res-hog}
\begin{tabular}{|l|c c c|c|l|c c c|}
\hline 
\multirow{2}{*}{Method} &
\multicolumn{3}{c|}{LibSVM + HOG (2000D)} &
\multicolumn{1}{c|}{\multirow{2}{*}{}}&
\multicolumn{1}{c|}{\multirow{2}{*}{Method}}&
\multicolumn{3}{c|}{MMD + HOG (2000D)} \\
\cline{2-4} \cline{7-9}
  & COIL100 & VOC2012 & Caltech101 & \multicolumn{1}{c|}{} & \multicolumn{1}{c|}{} & COIL100 & VOC2012 & Caltech101 \\
\hline \hline
ERE    & 0.0043 (~750)    & 0.4804 (2000)  & 0.3244 (1050)  &   & ERE   & 0.0071 (1300) & 0.5679 (2000)  & 0.3531 (~750)   \\
SPCA   & 0.0052 (2000)    & 0.4773 (1950)  & 0.3331 (~900)  &   & SPCA  & 0.0071 (2000) & 0.5339 (1600)  & 0.3228 (~850)   \\
RPCA   & 0.0057 (~700)    & 0.4669 (~550)  & 0.3068 (~650)  &   & RPCA  & 0.0071 (1600) & 0.5458 (1450)  & 0.3228 (~400)   \\
LRR    & 0.0067 (~200)    & 0.4681 (1100)  & 0.3102 (~200)  &   & LRR   & 0.0076 (~400) & 0.5080 (~800)  & 0.3108 (~250)   \\
FRR    & 0.0048 (~350)    & 0.4655 (1300)  & 0.3035 (~350)  &   & FRR   & 0.0110 (1000) & 0.5218 (~750)  & 0.3115 (~400)   \\
LLRR   & 0.0043 (~900)    & 0.4706 (~150)  & 0.2913 (~650)  &   & LLRR  & 0.0071 (1450) & 0.5118 (1550)  & 0.3024 (~400)   \\
SRRS   & 0.0043 (~600)    & 0.4624 (1300)  & 0.3026 (~550)  &   & SRRS  & 0.0095 (1300) & 0.5004 (~600)  & 0.3119 (~350)   \\
CLRR   & \bf{0.0036} (~300) & \bf{0.4491} (1500)  & \bf{0.2720} (~750)  &   & CLRR  & \bf{0.0058} (1400) & \bf{0.4776} (~100)  & \bf{0.2813} (~400)   \\
\hline 
\end{tabular}
\end{table*}
%
\begin{table*}[!t]
\centering
\caption{Classification error rates (Dimension) on testing data using SIFT feature.}
\label{tbl-res-sift}
\begin{tabular}{|l|c c c|c|l|c c c|}
\hline 
\multirow{2}{*}{Method} &
\multicolumn{3}{c|}{LibSVM + SIFT (2000D)} &
\multicolumn{1}{c|}{\multirow{2}{*}{}}&
\multicolumn{1}{c|}{\multirow{2}{*}{Method}}&
\multicolumn{3}{c|}{MMD + SIFT (2000D)} \\
\cline{2-4} \cline{7-9}
  & COIL100 & VOC2012 & Caltech101 & \multicolumn{1}{c|}{} & \multicolumn{1}{c|}{} & COIL100 & VOC2012 & Caltech101 \\
\hline \hline
ERE    & 0.0105 (1800)    & 0.4549 (2000)  & 0.2328 (~850)  &   & ERE   & 0.0067 (1950) & 0.5492 (2000)  & 0.2336 (1000)   \\
SPCA   & 0.0105 (1950)    & 0.4513 (1950)  & 0.2376 (2000)  &   & SPCA  & 0.0071 (2000) & 0.5269 (1600)  & 0.2340 (~650)   \\
RPCA   & 0.0081 (1400)    & 0.4369 (~450)  & 0.2291 (~500)  &   & RPCA  & 0.0076 (1700) & 0.5443 (1150)  & 0.2246 (~550)   \\
LRR    & 0.0090 (~250)    & 0.4460 (1000)  & 0.2352 (~300)  &   & LRR   & 0.0110 (~300) & 0.5333 (~600)  & 0.2494 (~250)   \\
FRR    & 0.0057 (~900)    & 0.4431 (1900)  & 0.2383 (~500)  &   & FRR   & 0.0095 (1250) & 0.5265 (~850)  & 0.2395 (~350)   \\
LLRR   & 0.0074 (2000)    & 0.4358 (~550)  & 0.2197 (~700)  &   & LLRR  & 0.0067 (2000) & 0.5093 (2000)  & 0.2243 (~600)   \\
SRRS   & 0.0090 (1200)    & 0.4384 (~600)  & 0.2230 (~900)  &   & SRRS  & 0.0079 (1600) & 0.4951 (~800)  & 0.2219 (~750)   \\
CLRR   & \bf{0.0052} (1000) & \bf{0.4230} (~300)  & \bf{0.2109} (1000)  &  & CLRR  & \bf{0.0043} (1400) & \bf{0.4659} (~250)  & \bf{0.2102} (~500)   \\
\hline 
\end{tabular}
\end{table*}

We adopt the classification error rate as evaluation criterion. Though SRRS simply used Nearest-Neighbor (NN) as the classifier, we find it ineffective on large image databases. Instead, Support Vector Machine (SVM) \cite{bishop-book2006-prml} has been proved to be powerful in many applications, and here we utilized LibSVM \cite{chang-tist2011-libsvm} as primary classifier. In particular, we used the $C$-SVC model \cite{chang-tist2011-libsvm} with linear kernel and chose the best $C$ from $2^{[-3:1:6]}$ by two-fold cross validation on the validation data. During parameter selection for all compared methods, $C$ was kept to 10. To better convince merits of the proposed method, we have also reported the results using Minimum Mahalonobis Distance (MMD) classifier \cite{mather2009classification}, which leverages within class scatter matrix of the data.

To sufficiently explore the performances of the above methods, we did some tests on the four different types of features, \ie, GIST (512D), LBP (1239D), HOG (2000D), SIFT (2000D), separately. The former two use the original dimension while the latter ones use the reduced dimension 2000 by PCA \cite{bishop-book2006-prml}. The results of them are reported in Tables~\ref{tbl-res-gist}, \ref{tbl-res-lbp}, \ref{tbl-res-hog} and \ref{tbl-res-sift}, respectively, where each record is paired with the dimensionality which produces the result. The best record in each column of the table is highlighted in boldface.

%
\begin{table*}[!t]
\centering
\caption{Classification error rates on testing data using combined features with 400, 800 and 1600 dimensions by LibSVM.}
\label{tbl-res-combine}
\begin{tabular}{|l|c c c|c| c c c|c|c c|}
\hline 
\multirow{2}{*}{Method} &
\multicolumn{3}{c|}{Combined Features (400D)} &
\multicolumn{1}{c|}{\multirow{2}{*}{}}&
\multicolumn{3}{c|}{Combined Features (800D)} &
\multicolumn{1}{c|}{\multirow{2}{*}{}}&
\multicolumn{2}{c|}{Combined Features (1600D)} \\
\cline{2-4} \cline{6-8} \cline{10-11}
  & COIL100 & VOC2012 & Caltech101 & \multicolumn{1}{c|}{} & COIL100 & VOC2012 & Caltech101 & \multicolumn{1}{c|}{} & VOC2012 & Caltech101 \\
\hline \hline
ERE   & 0.0057 (350) & 0.4559 (350) & 0.3167 (200)  & & 0.0081 (650) & 0.4419 (750) & 0.2845 (400) & & 0.4413 (1600) & 0.2609 (~750)  \\
SPCA  & 0.0062 (250) & 0.4561 (350) & 0.3130 (400)  & & 0.0081 (650) & 0.4421 (650) & 0.2912 (650) & & 0.4398 (1250) & 0.2786 (1450)  \\
RPCA  & 0.0048 (200) & 0.4390 (400) & 0.2874 (200)  & & 0.0052 (350) & 0.4402 (750) & 0.2732 (400) & & 0.4386 (1600) & 0.2548 (~700)  \\
LRR   & 0.0057 (350) & 0.4482 (400) & 0.2929 (200)  & & 0.0048 (400) & 0.4397 (750) & 0.2650 (400) & & 0.4477 (1400) & 0.2485 (~750)  \\
FRR   & 0.0057 (350) & 0.4352 (400) & 0.2929 (200)  & & 0.0043 (350) & 0.4340 (750) & 0.2657 (350) & & 0.4391 (1500) & 0.2532 (~800)  \\
LLRR  & 0.0052 (350) & \bf{0.4263} (350) & 0.2707 (200)  & & 0.0052 (400) & 0.4223 (750) &0.2492 (400) & & 0.4329 (1400) & 0.2351 (~800)  \\
SRRS  & 0.0057 (400) & 0.4311 (350) & 0.2765 (200)  & & 0.0048 (400) & 0.4309 (750) & 0.2593 (350) & & 0.4381 (1400) & 0.2524 (~750)  \\
CLRR &\bf{0.0052} (400)&  0.4297 (250)&\bf{0.2472} (200) & & \bf{0.0041} (750)& \bf{0.4221} (750)&\bf{0.2346} (400)&&\bf{0.4186} (1300)&\bf{0.2274} (~650)\\
\hline 
\end{tabular}
\end{table*}
%
\begin{table*}[!t]
\centering
\caption{Classification error rates on testing data for examining individual components of the proposed CLRR method.}
\label{tbl-clrr-components}
\begin{tabular}{|l|c c|c c|c|c c|c c|}
\hline 
\multirow{3}{*}{Method} &
\multicolumn{4}{c|}{GIST (512D)} &
\multicolumn{1}{c|}{\multirow{3}{*}{}}&
\multicolumn{4}{c|}{Combined Features (1600D)} \\
\cline{2-5} \cline{7-10}
 & \multicolumn{2}{c|}{LibSVM} & \multicolumn{2}{c|}{MMD}   & \multicolumn{1}{c|}{}   & \multicolumn{2}{c|}{LibSVM} & \multicolumn{2}{c|}{MMD} \\
\cline{2-5} \cline{7-10}
 & VOC2012 & Caltech101  & VOC2012 & Caltech101 & \multicolumn{1}{c|}{} & VOC2012 & Caltech101 & VOC2012 & Caltech101 \\
\hline \hline
CLRR      & 0.4912 (350) & 0.2816 (500) & 0.5936 (200) & 0.3235 (100)  & & 0.4186 (1300) & 0.2374 (~650) & 0.5061 (1050) & 0.2170 (~550) \\
CLRR$_a$  & 0.5163 (400) & 0.3038 (300) & 0.6132 (150) & 0.3425 (100)  & & 0.4301 (1150) & 0.2525 (1000) & 0.5239 (1100) & 0.2305 (1200) \\
CLRR$_b$  & 0.4971 (300) & 0.2990 (500) & 0.6010 (300) & 0.3390 (~50)  & & 0.4260 (~750) & 0.2408 (~900) & 0.5152 (~900) & 0.2261 (~600) \\
CLRR$_c$  & 0.5206 (350) & 0.2911 (400) & 0.6367 (200) & 0.3418 (200)  & & 0.4457 (1000) & 0.2450 (~300) & 0.5411 (1100) & 0.2376 (~550) \\
\hline 
\end{tabular}
\end{table*}
%

From these results, a number of interesting observations can be found in the following.
\begin{itemize}
\item CLRR systematically and consistently outperforms the compared methods on image classification. We attribute this to the fact that our method fully takes advantage of the discriminant information encoded by the Laplacian regularizer and the supervised information by the adaptive least squares regularizer. Besides, enforcing each column sum of the low-rank matrix $\mathbf{Z}$ to be one has positive effects on classification.

\item The classification error rates of SRRS and CLRR are lower than those of other low-rank methods on GIST and HOG features of VOC2012, which demonstrates that supervised constraints are indeed beneficial for classification. Futhermore, CLRR achieves more performance improvements than that of SRRS, which empirically validates the efficacy of the enforced least squares regularizer.

\item LLRR can produce better classification results than most other methods on VOC2012 and Caltech101, which indicates the advantages of constructing the dictionary using both observed data and hidden data to complement the insufficient observations. Thus, it might be helpful for CLRR by considering this in a possible way.

\item ERE and SPCA, though simple, performs satisfactorily on COIL100, even better than some low-rank based methods. In particular, SPCA enjoys more promising performances than ERE overall, which suggests that considering class-specific information by assigning different weights indeed leads to a better model. However, it seems difficult for them to perform well on VOC2012 and Caltech101, which are more challenging than COIL100.

\item For features using original dimension, GIST exhibits better performances than LBP using LibSVM while LBP produces better results than GIST on COIL100 and VOC2012 using MMD, which indicates more dimensions do not necessarily enhance the performance. For features using reduced dimension, SIFT performs better than HOG on VOC2012 and Caltech101, which suggests SIFT would be a better choice for challenging databases. Overall, HOG and SIFT yield much better results than GIST and LBP in most cases, which demonstrates that reducing dimensions by PCA enables boosting classification performances.

\item Generally, the results derived from LibSVM are superior to those of MMD, which is more obvious on the latter two databases. This reflects that SVM often performs better than MMD in real-world applications. However, MMD could obtain comparable or even better results on COIL100, which means both of them can be used to well handle simple databases, where it is easy to distinguish different categories covered in the data.

\end{itemize}
%
\begin{figure*}[!t]
\centering
\includegraphics[width=0.19\textwidth]{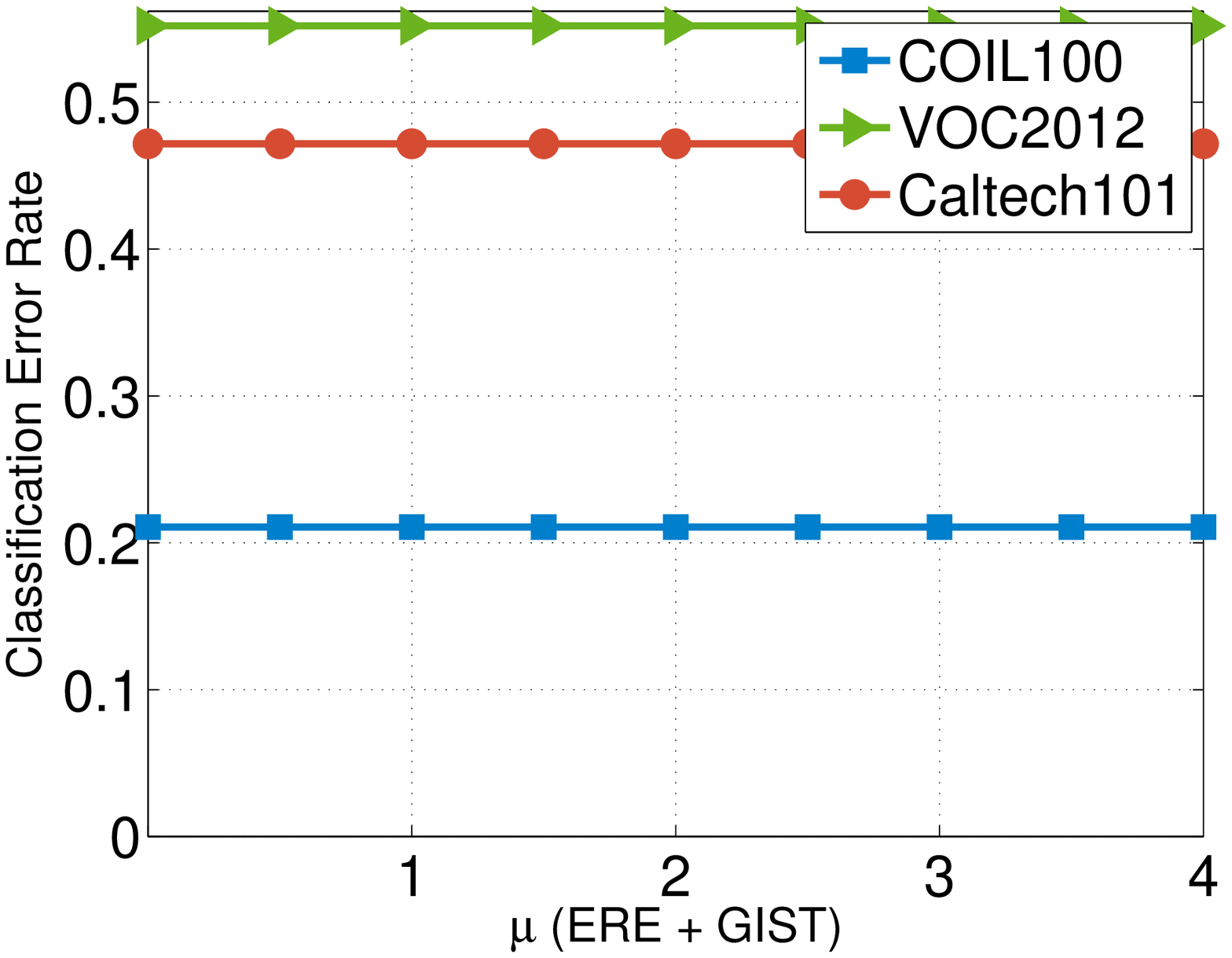}
\includegraphics[width=0.19\textwidth]{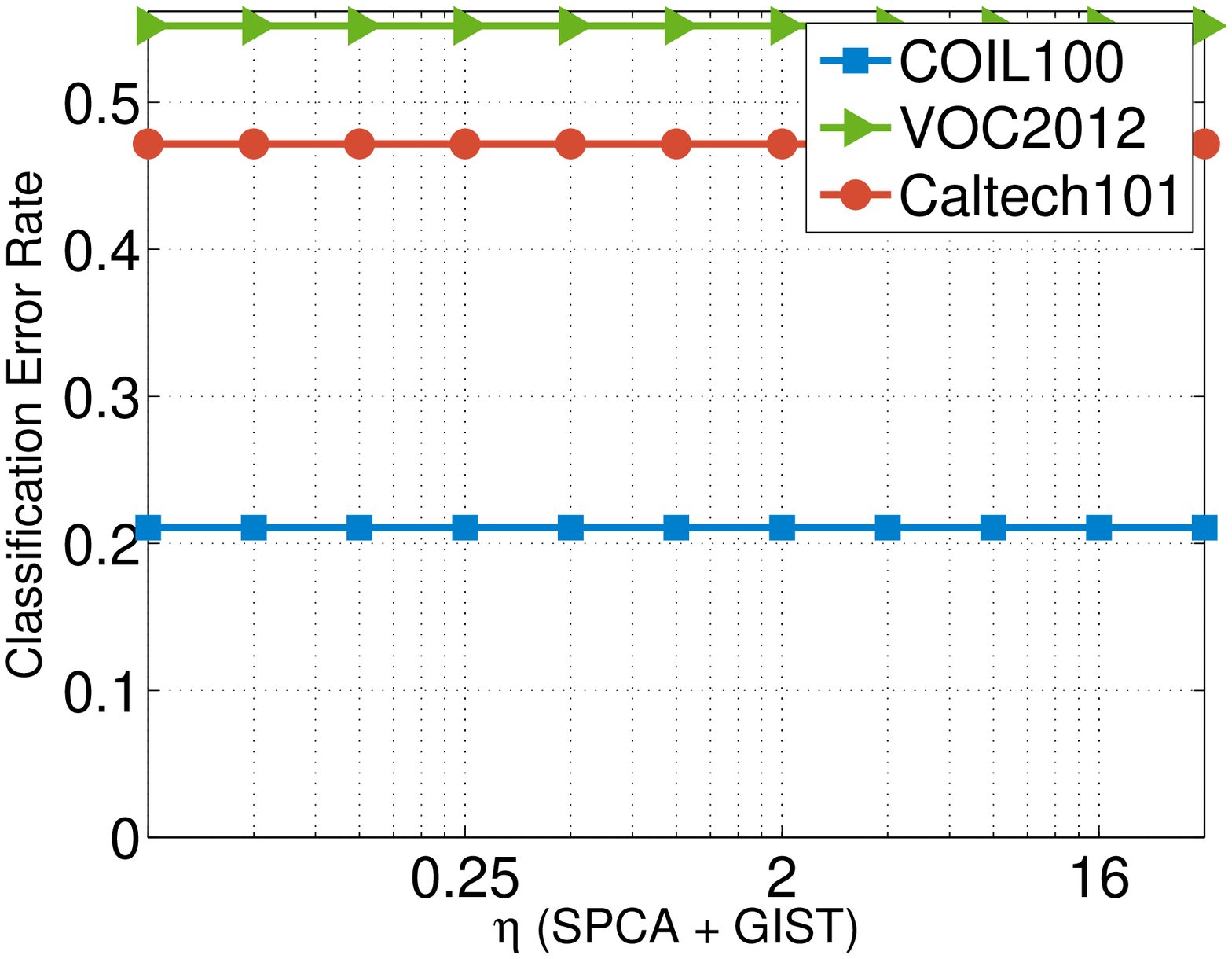}
\includegraphics[width=0.19\textwidth]{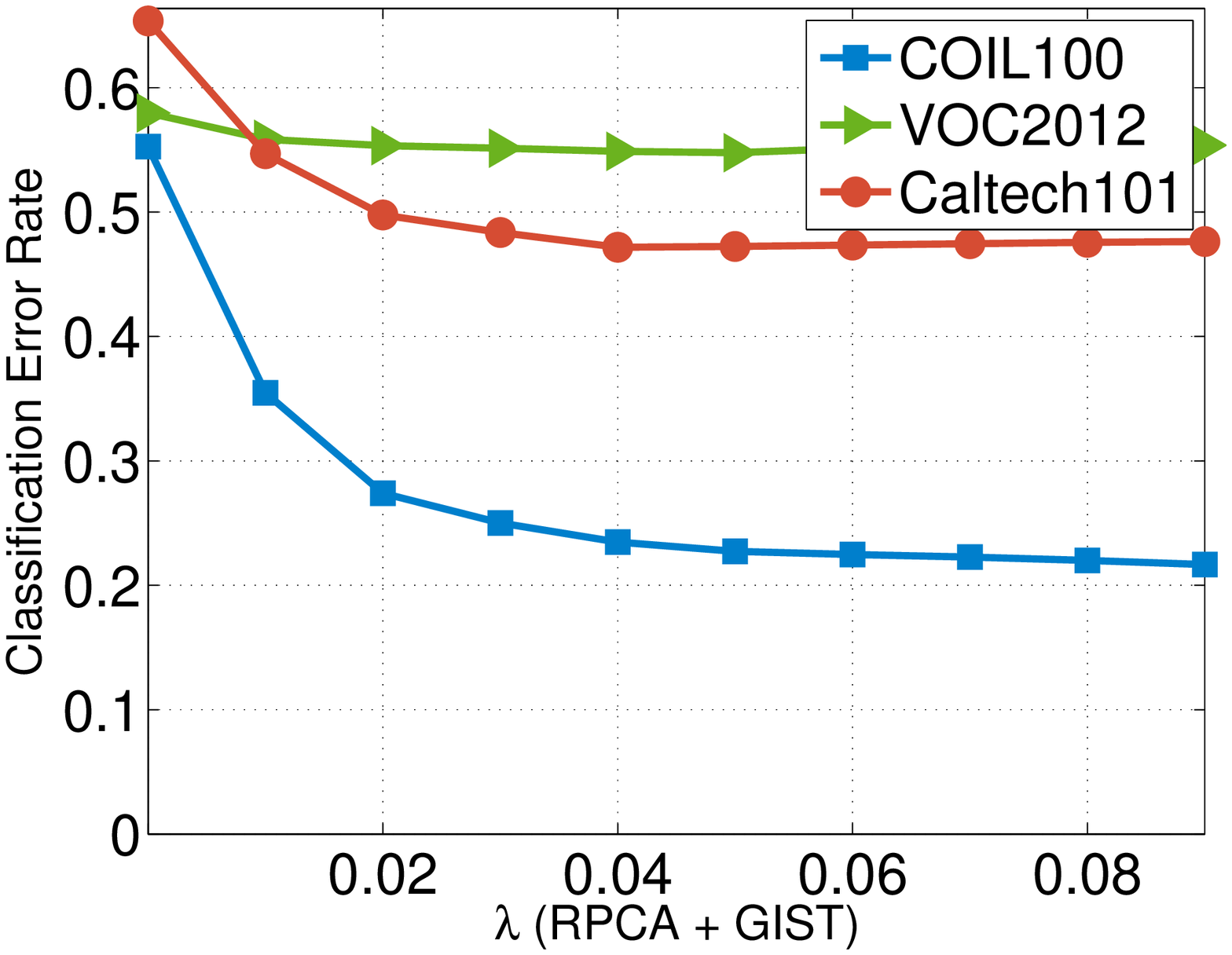}
\includegraphics[width=0.19\textwidth]{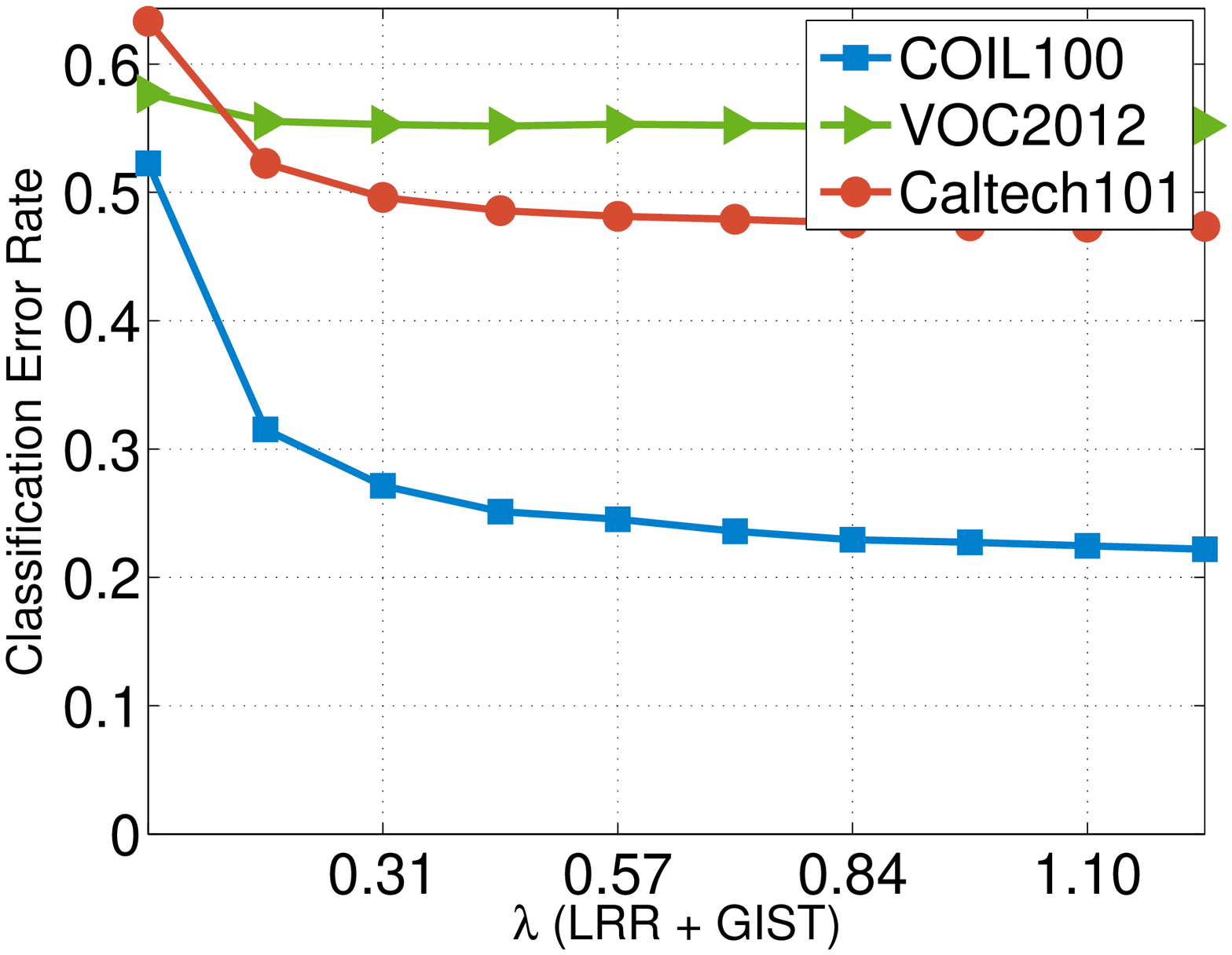}
\includegraphics[width=0.19\textwidth]{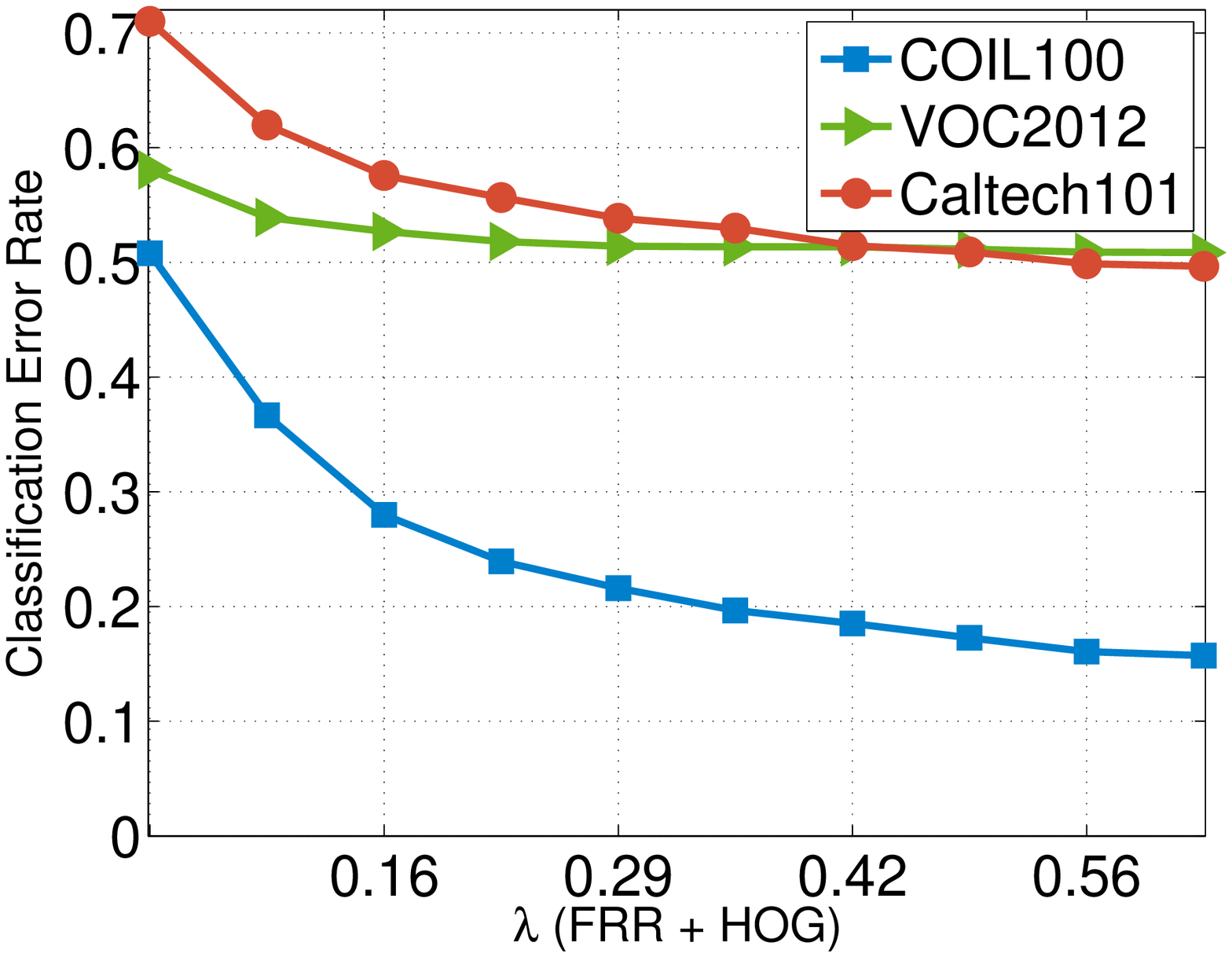}

\includegraphics[width=0.19\textwidth]{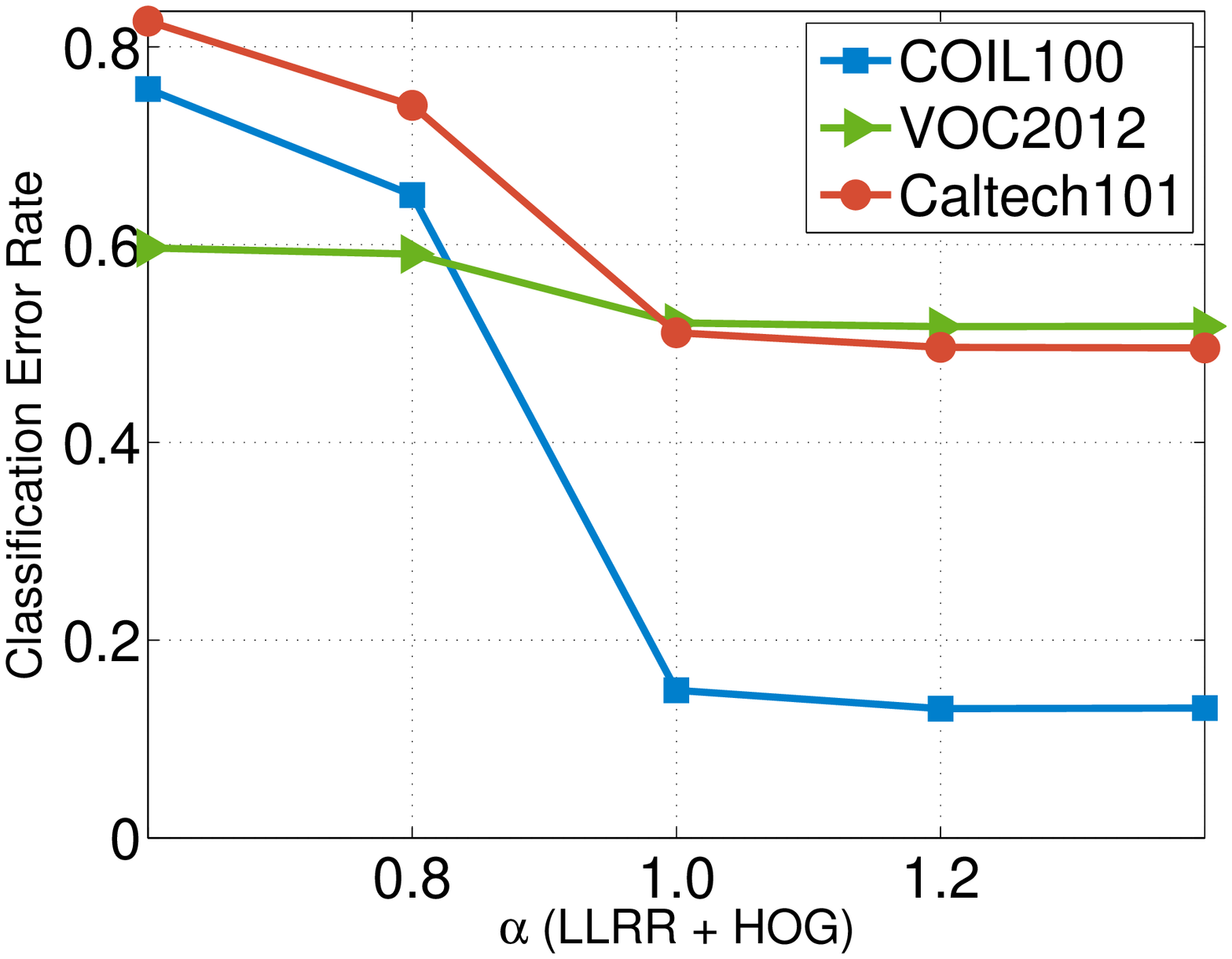}
\includegraphics[width=0.19\textwidth]{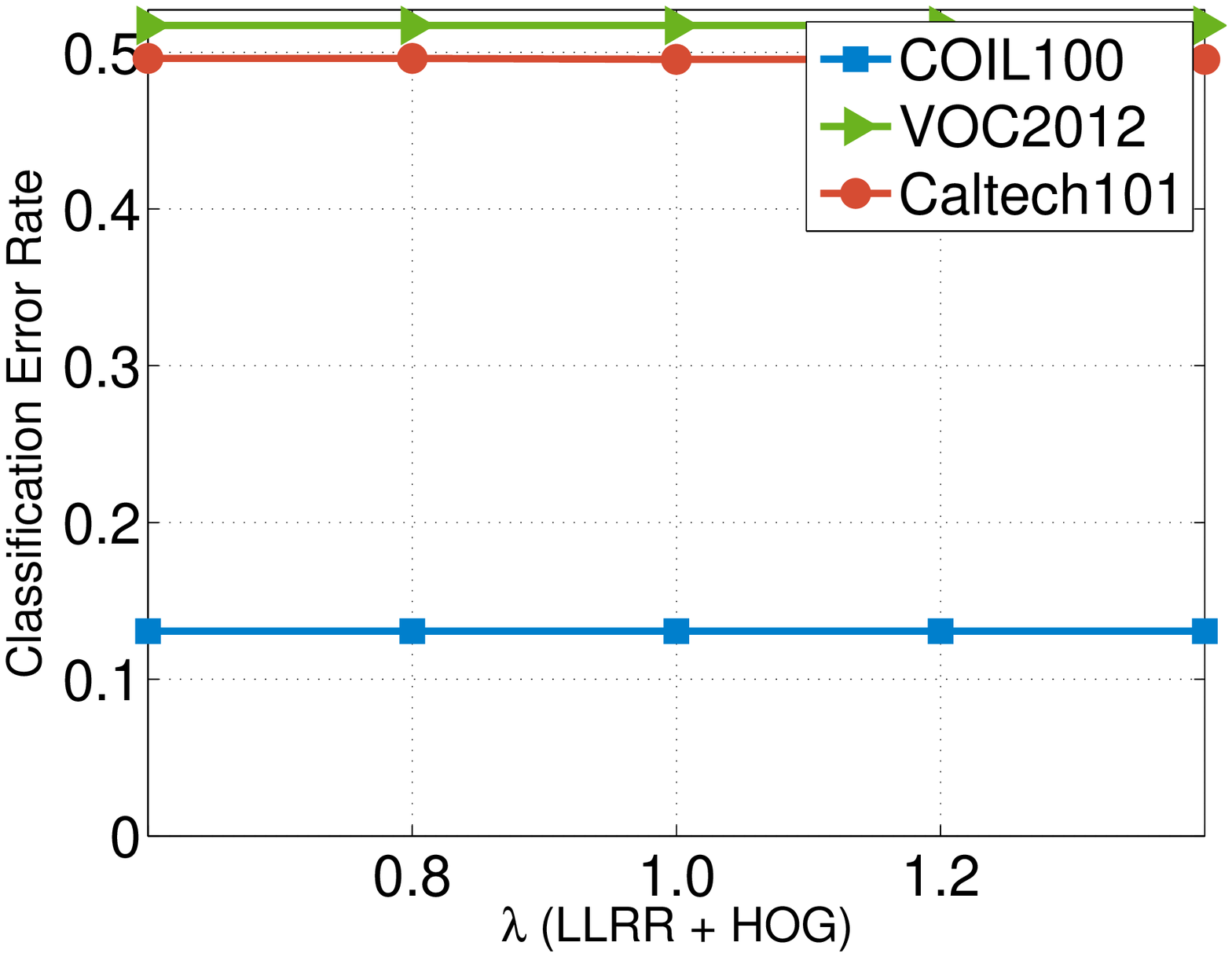}
\includegraphics[width=0.19\textwidth]{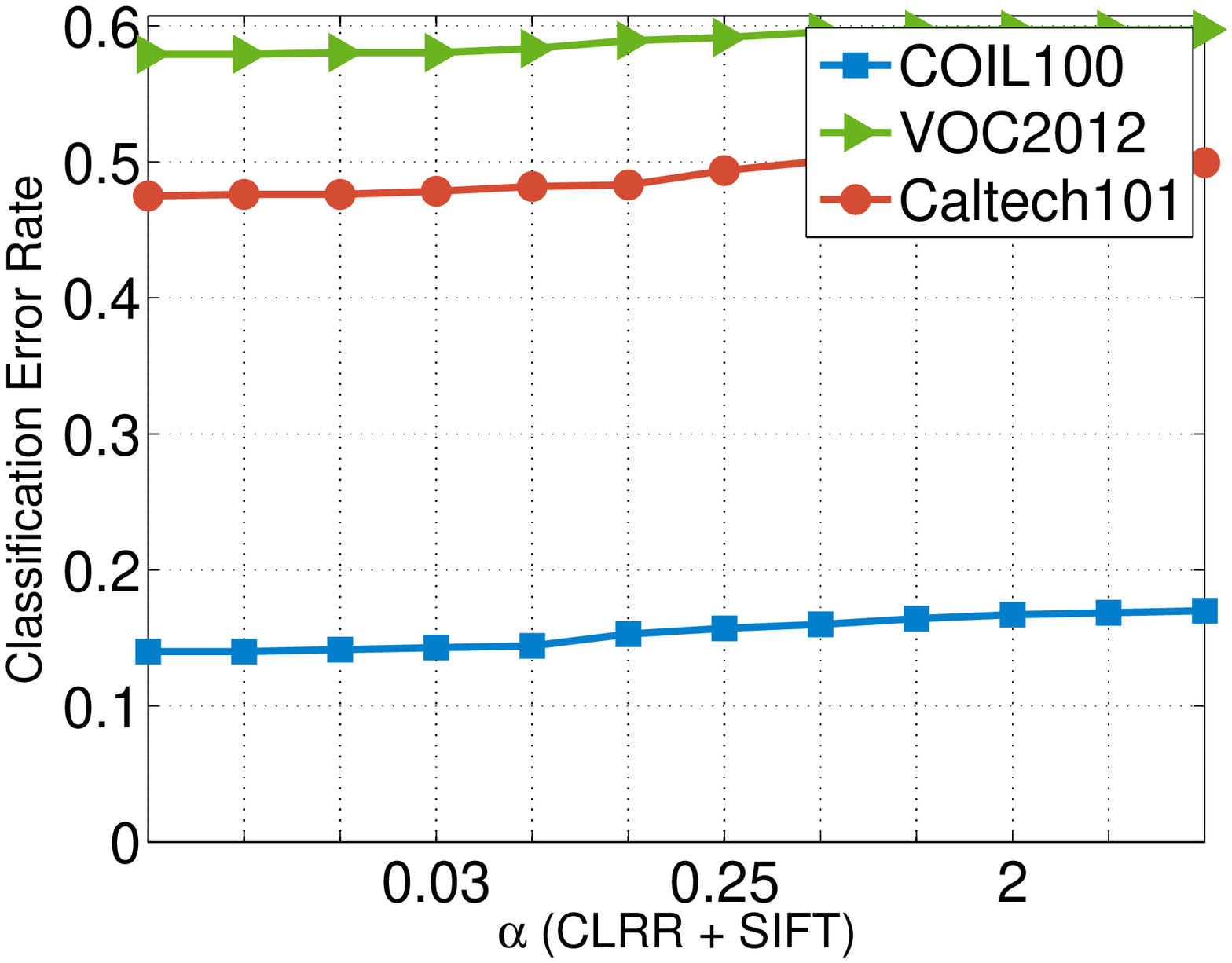}
\includegraphics[width=0.19\textwidth]{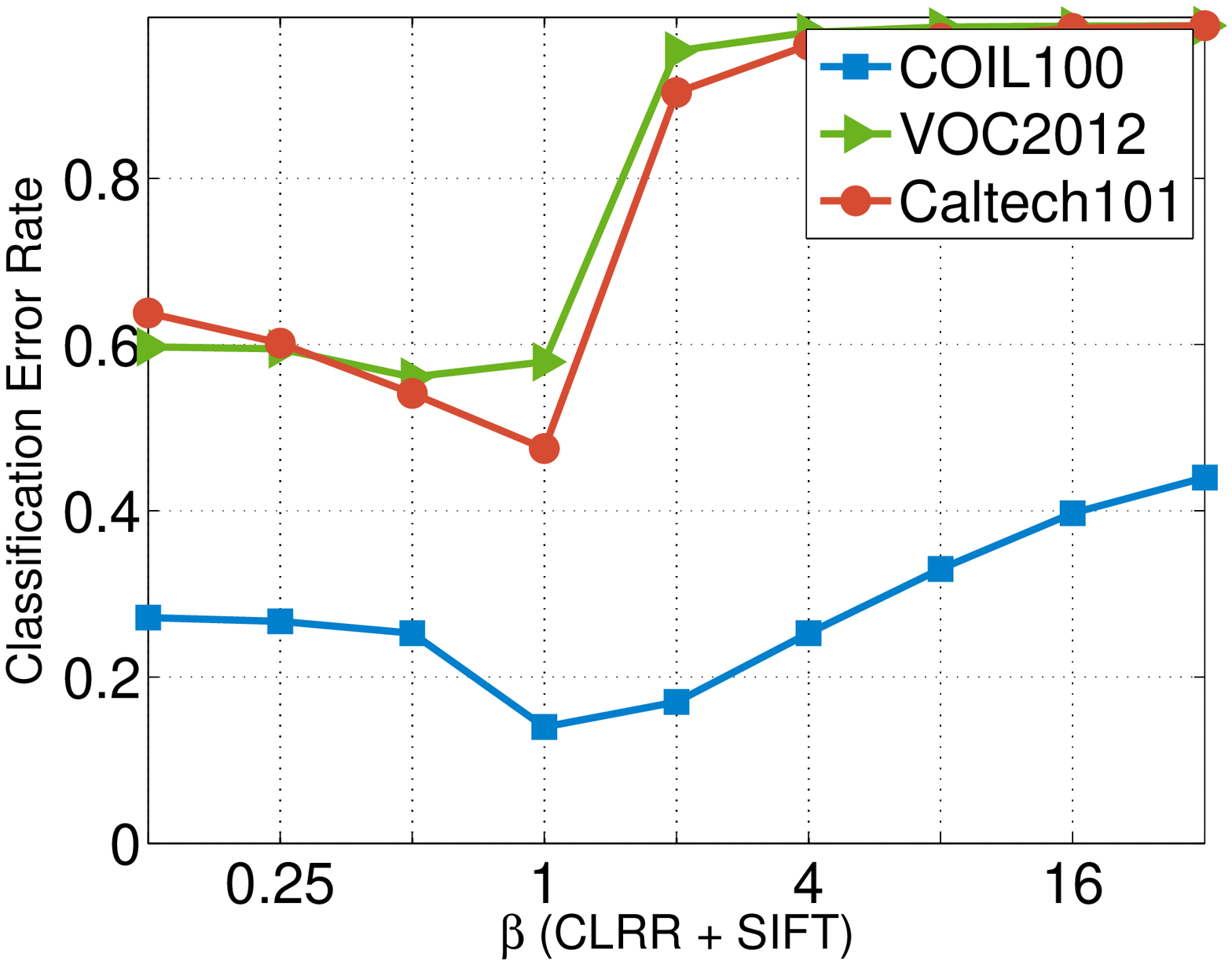}
\includegraphics[width=0.19\textwidth]{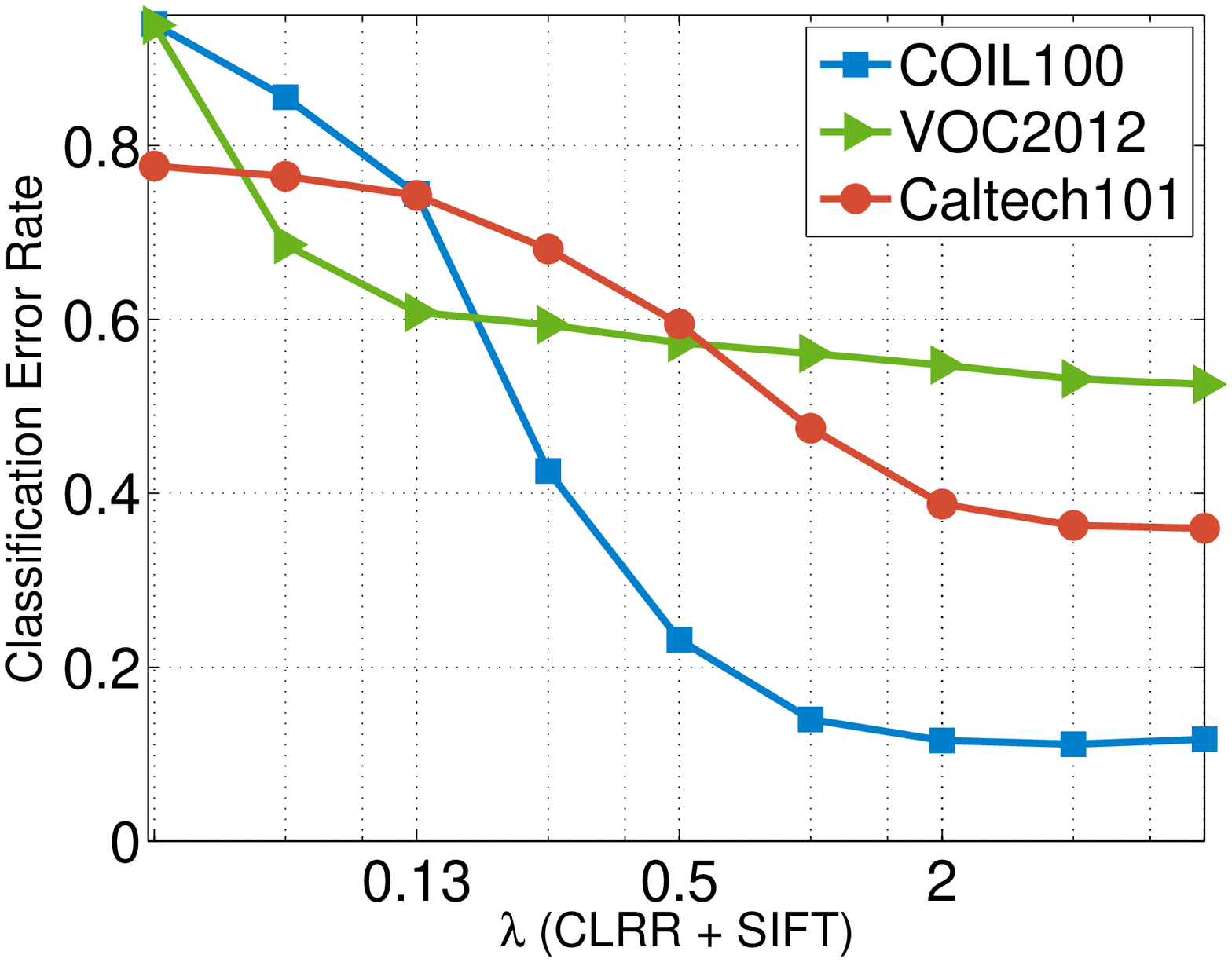}
\caption{Parameter selection on validation data of three image databases. The parameter $C$ of LibSVM was set to 10, and others perform similarly.}
\label{fig-methodParaSel}
\end{figure*}

In order to test the cumulative effect of combined features with different dimensions, we integrated the four kinds of features for each database. In particular, the dimension of each feature was first reduced to 100, 200, 400 by PCA individually, and then different features were concatenated to generate one vector for each sample. Consequently, we can obtain the final 400, 800, 1600-dimensional data vectors for testing. The results produced by LibSVM are shown in Table~\ref{tbl-res-combine}, where the best results are highlighted in boldface. From these results, it can be observed that higher dimension only shows slight improvements but requiring more computations. Actually, the cumulative effects are insignificant compared to the results produced by separated SIFT feature in Table~\ref{tbl-res-sift}, which indicates that sometimes one informative feature is good enough for classification task.

To examine the individual components of CLRR, we compare it with several alternatives of its objective function in Eq.~(\ref{eq:clrr-2}), \ie, we use CLRR$_a$ to denote the method removing the least squares term by setting $\beta=0$, CLRR$_b$ removing the Laplacian term by setting $\alpha=0$, and CLRR$_c$ removing the low-rank term $\|\mathbf{Z}\|_*$ as well as setting $\beta=1$. Previous parameter selections regarding CLRR also apply to these alternatives. For the variant CLRR${_c}$, we directly used the projection matrix $\mathbf{P}$ to map both the training data and the testing data into the new subspace, since the low-rank term was removed in this situation, failing to capture the coefficient matrix $\mathbf{Z}$ for the testing data due to the lack of label information. Unlike previous variants degenerating the objective function to LRR by abandoning the constraints, we could not optimize $\mathbf{Z}$ on the testing data since it is not included in the objective function. The tests were carried out on one separated feature, \ie, GIST (512D), and one combined feature with 1600 dimensions, respectively, the results of which on VOC2012 and Caltech101 are recorded in Table~\ref{tbl-clrr-components}. From these results and those in Table~\ref{tbl-res-gist} and \ref{tbl-res-combine}, we see that CLRR$_{a}$ and CLRR$_{b}$ are comparable and even better than other methods, which indicates that using Laplacian regularizer or least squares regularizer individually is still able to capture informative representation. Moreover, CLRR$_{b}$ can reduce the error rate in a slightly larger magnitude than CLRR$_{a}$, which suggests explicit label structure may have stronger guidance on learning low-rank subspaces than between-class scatter based Laplacian regularizer. More importantly, coupling both of the two supervised regularizers gives satisfying improvements on classification performance. In addition, CLRR$_{c}$ performs well on Caltech101 while it reports worse records on VOC2012, which would be due to the reason that the learned data representation of VOC2012 only has 20 dimensions, and thus some reliable information would miss during the projection.

%
\begin{figure*}[!t]
\centering
\includegraphics[width=0.19\textwidth]{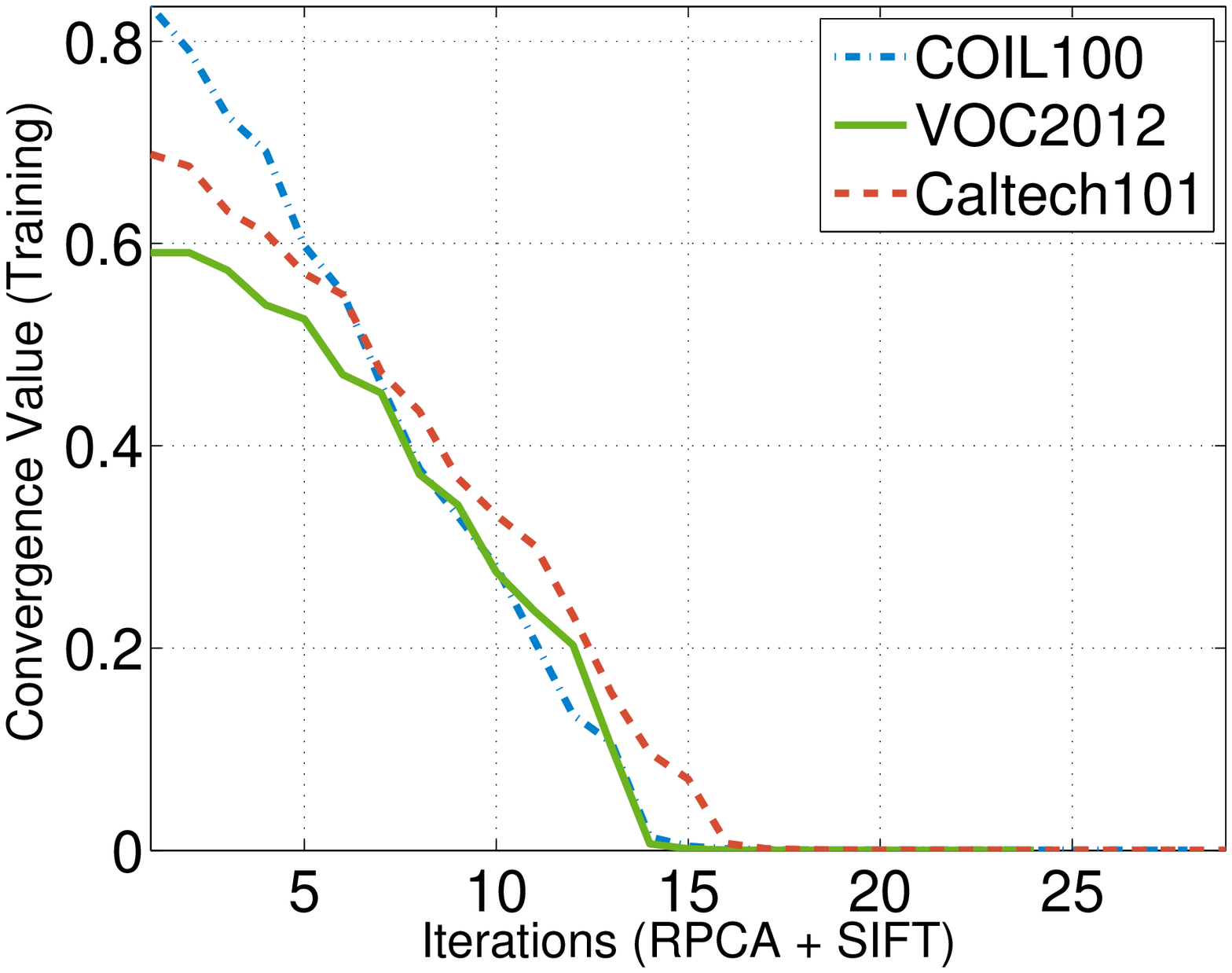}
\includegraphics[width=0.19\textwidth]{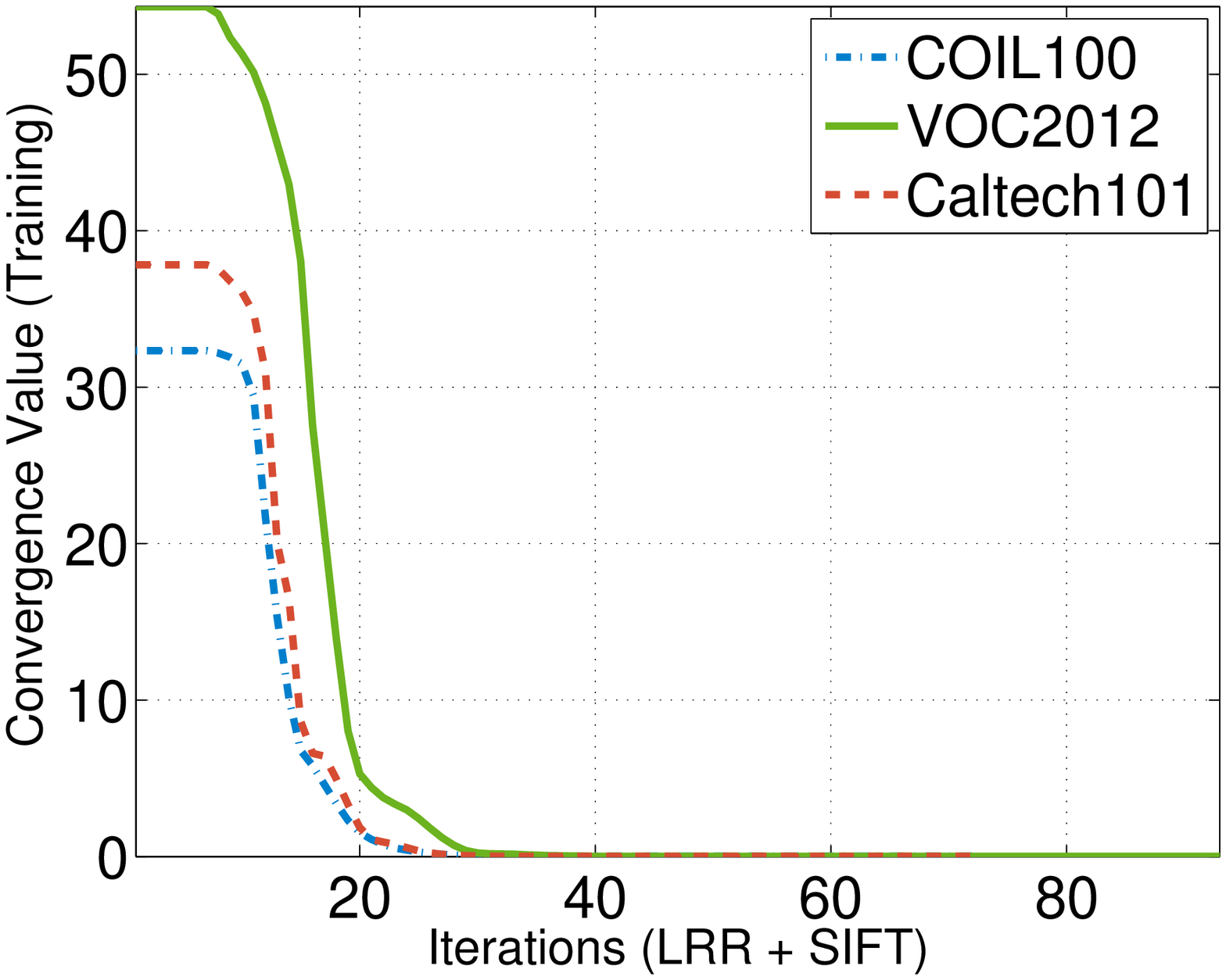}
\includegraphics[width=0.19\textwidth]{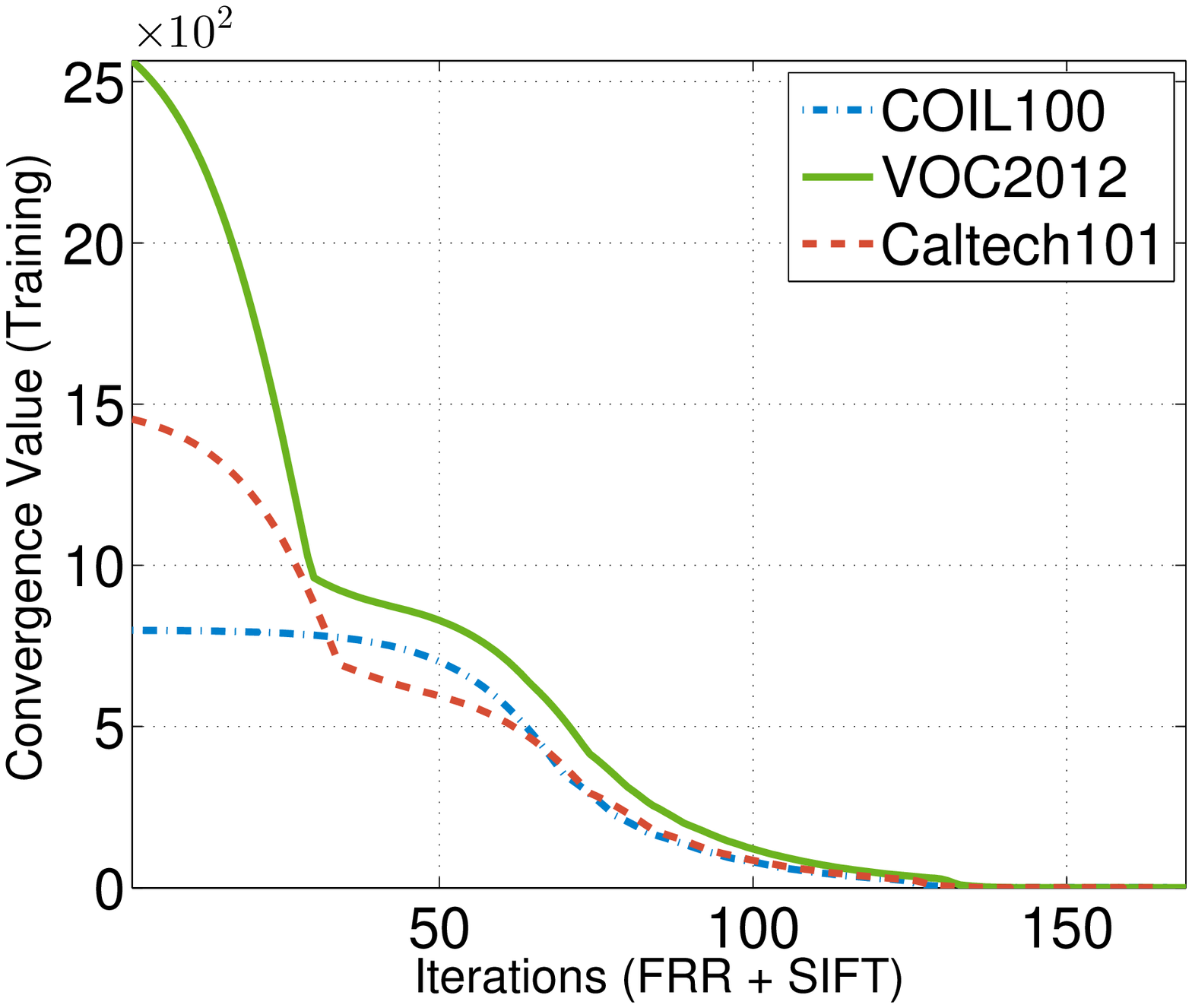}
\includegraphics[width=0.19\textwidth]{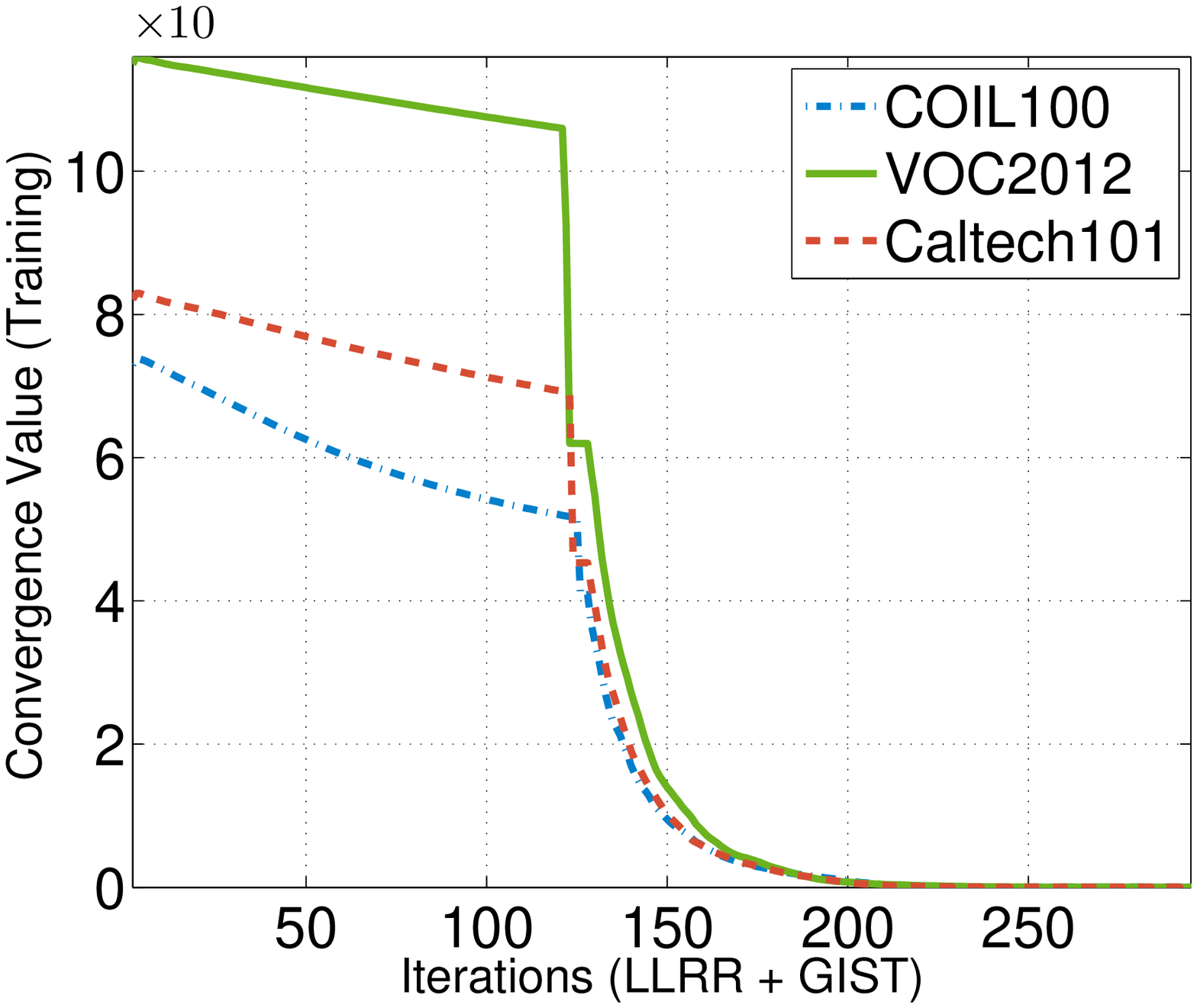}
\includegraphics[width=0.19\textwidth]{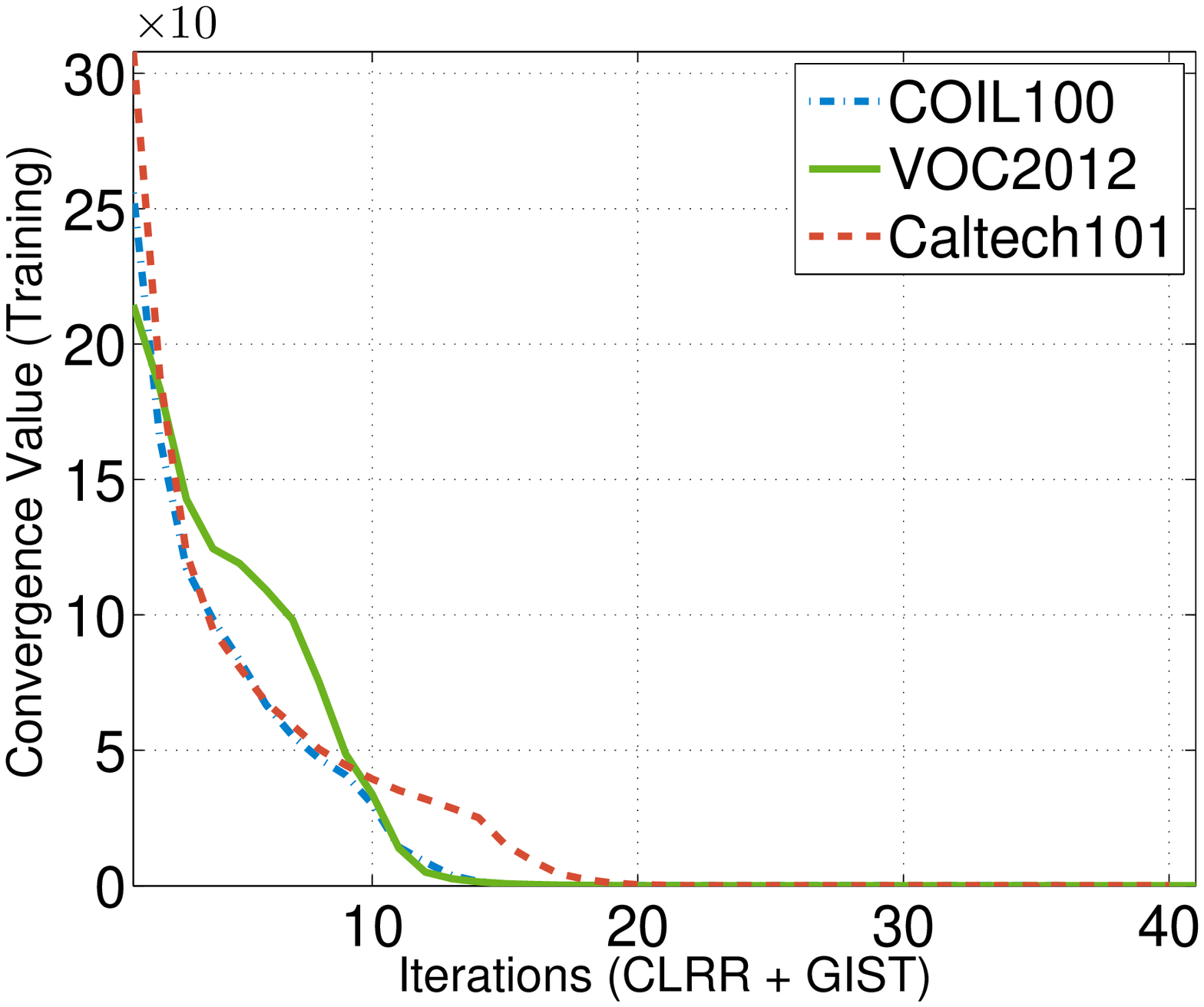}
\caption{Convergence curves on training data of three image databases. Others have similar convergence curves.}
\label{fig-converge}
\end{figure*}
%
%
\begin{figure*}[!t]
\centering
\includegraphics[width=0.19\textwidth]{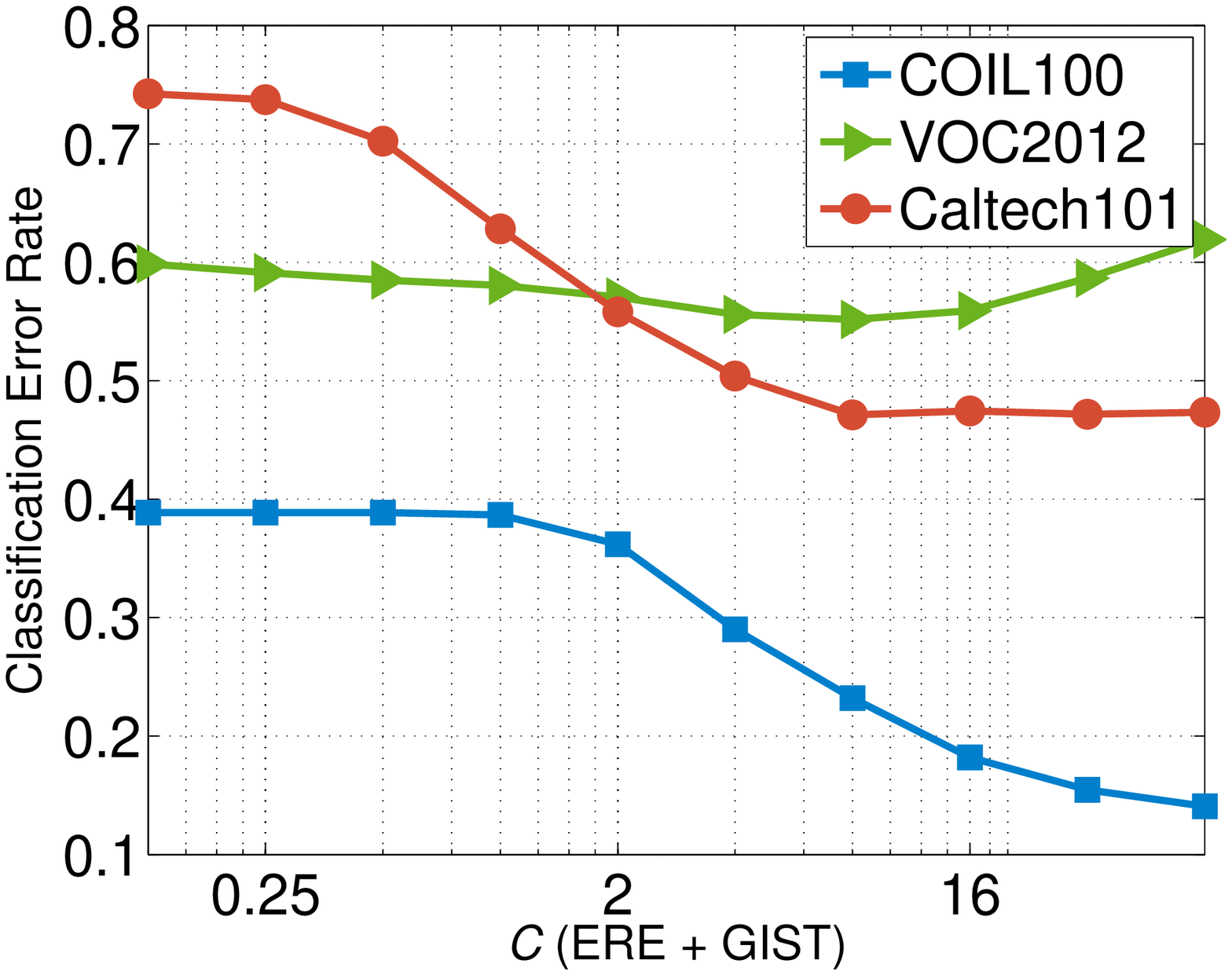}
\includegraphics[width=0.19\textwidth]{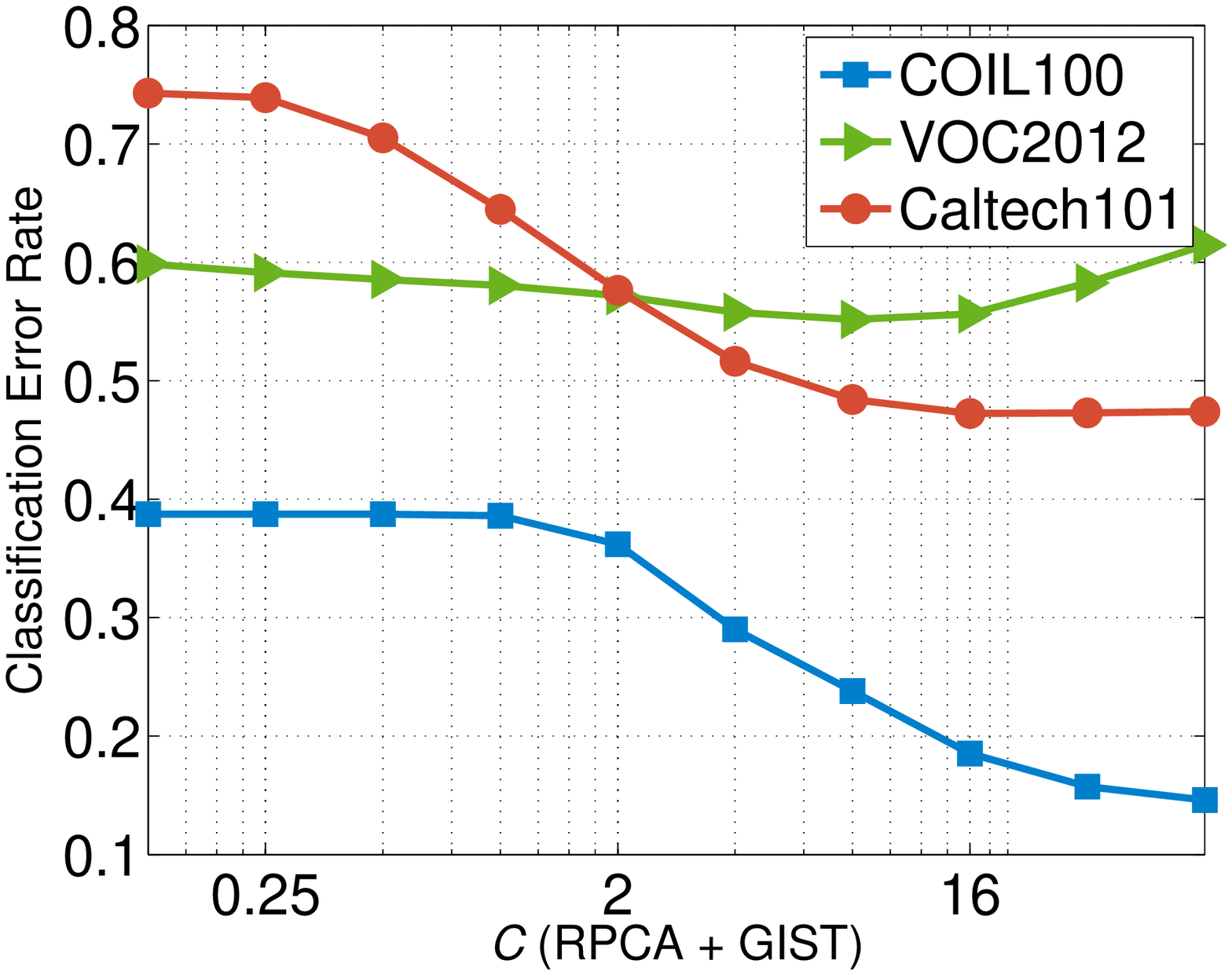}
\includegraphics[width=0.19\textwidth]{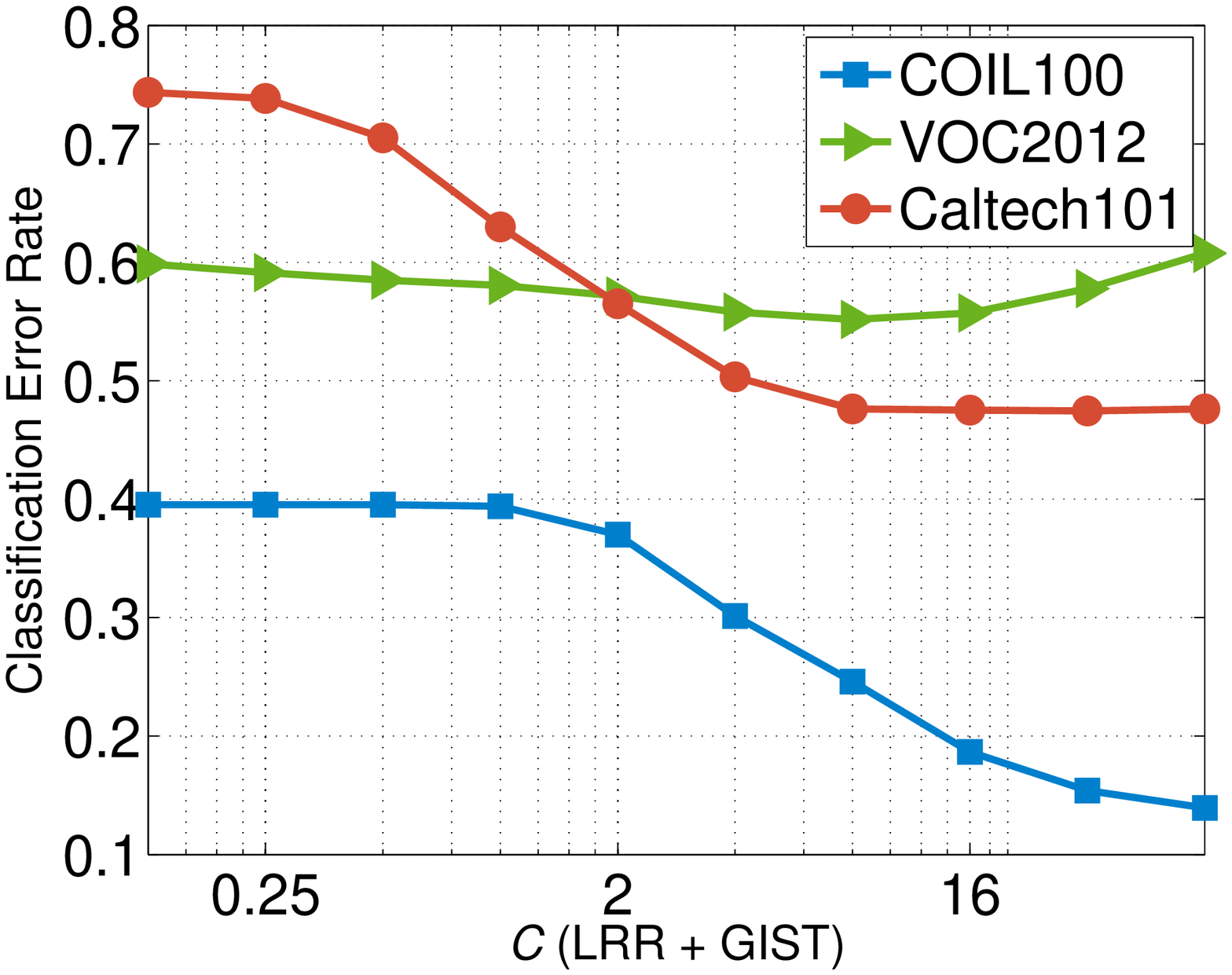}
\includegraphics[width=0.19\textwidth]{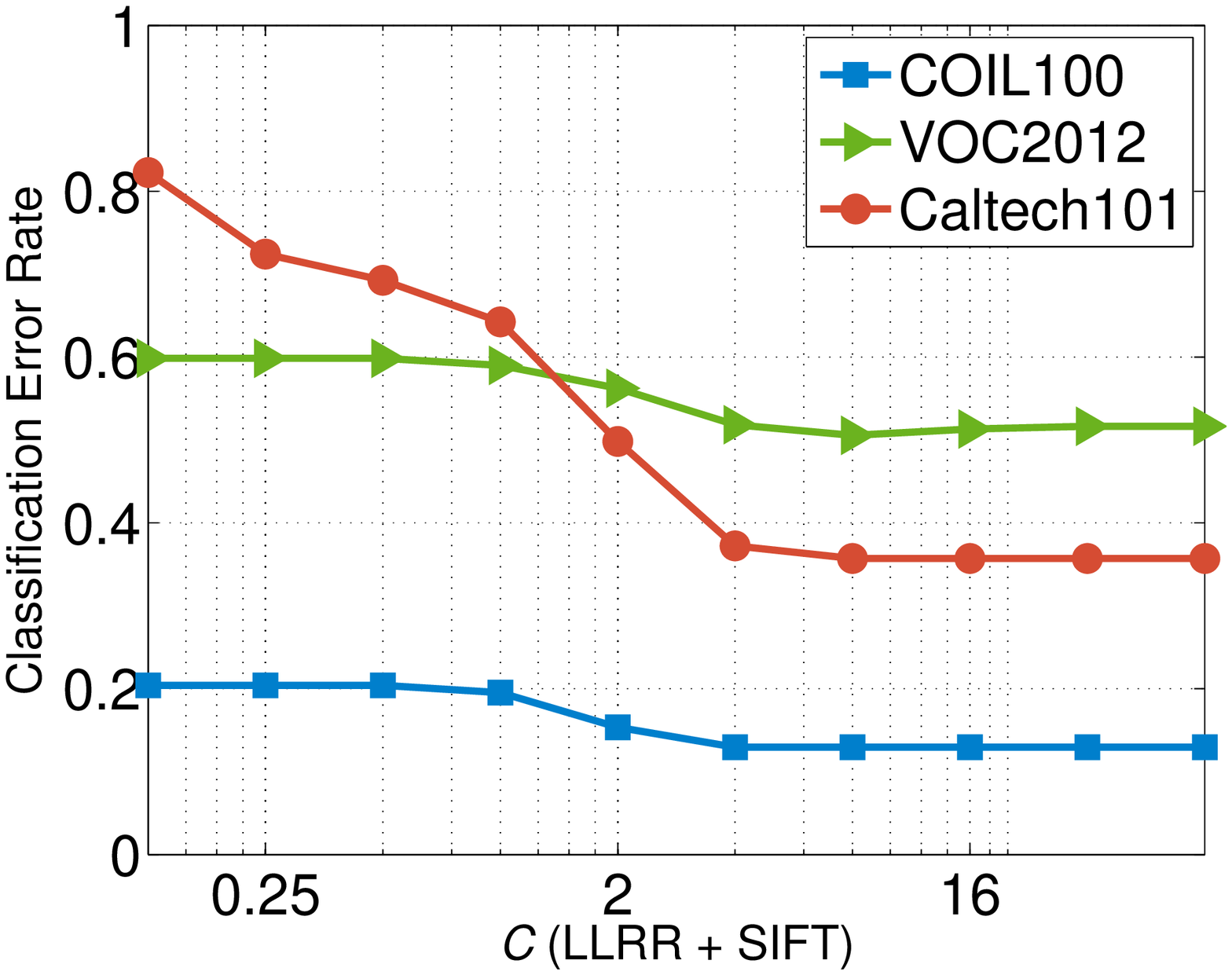}
\includegraphics[width=0.19\textwidth]{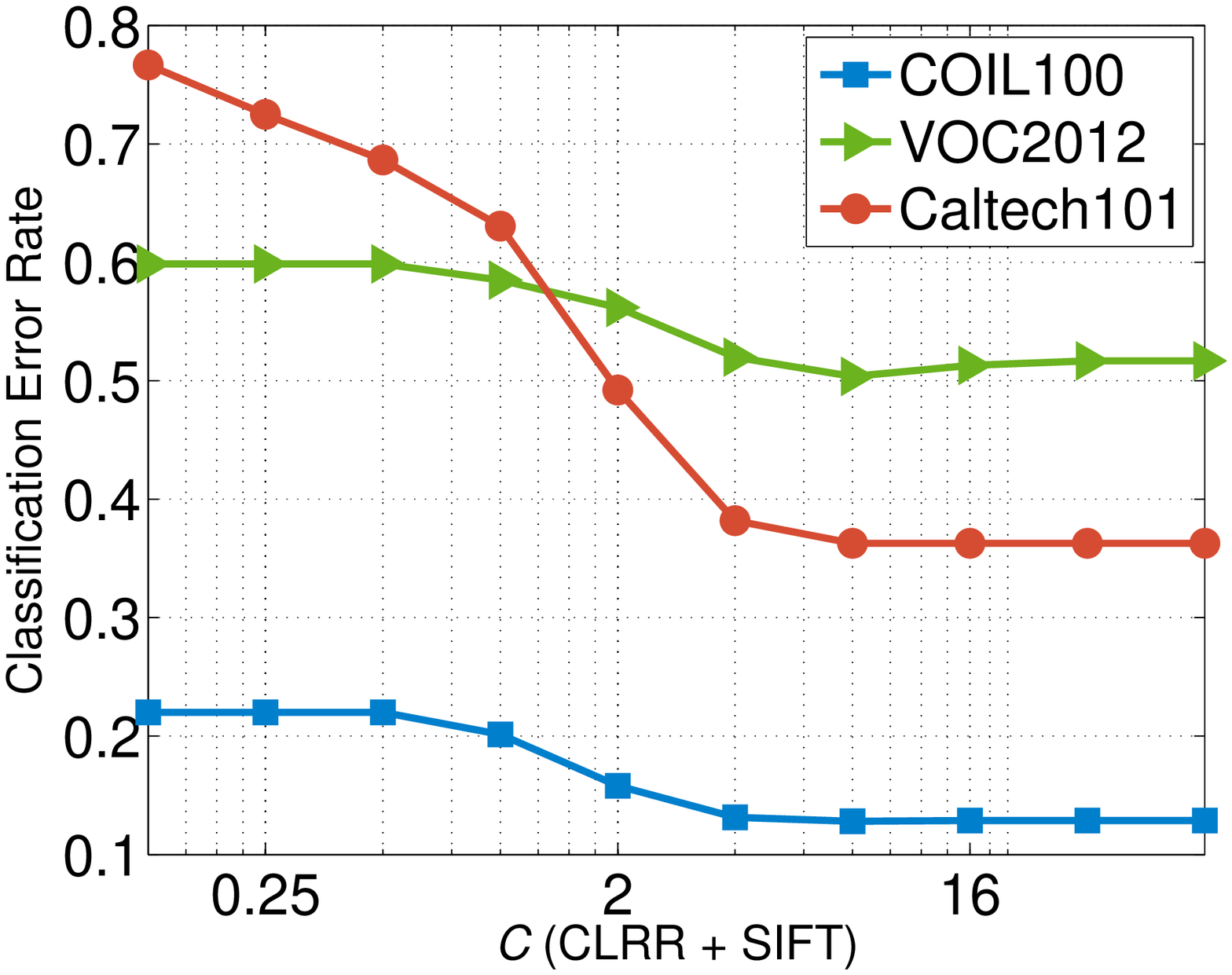}
\caption{Model selection of parameter $C$ in the $C$-SVC model of LibSVM on validation set of three image databases. Others follow the same fashion.}
\label{fig-svmParaSel}
\end{figure*}
%
%
\begin{figure*}[!t]
\centering
\includegraphics[width=0.19\textwidth]{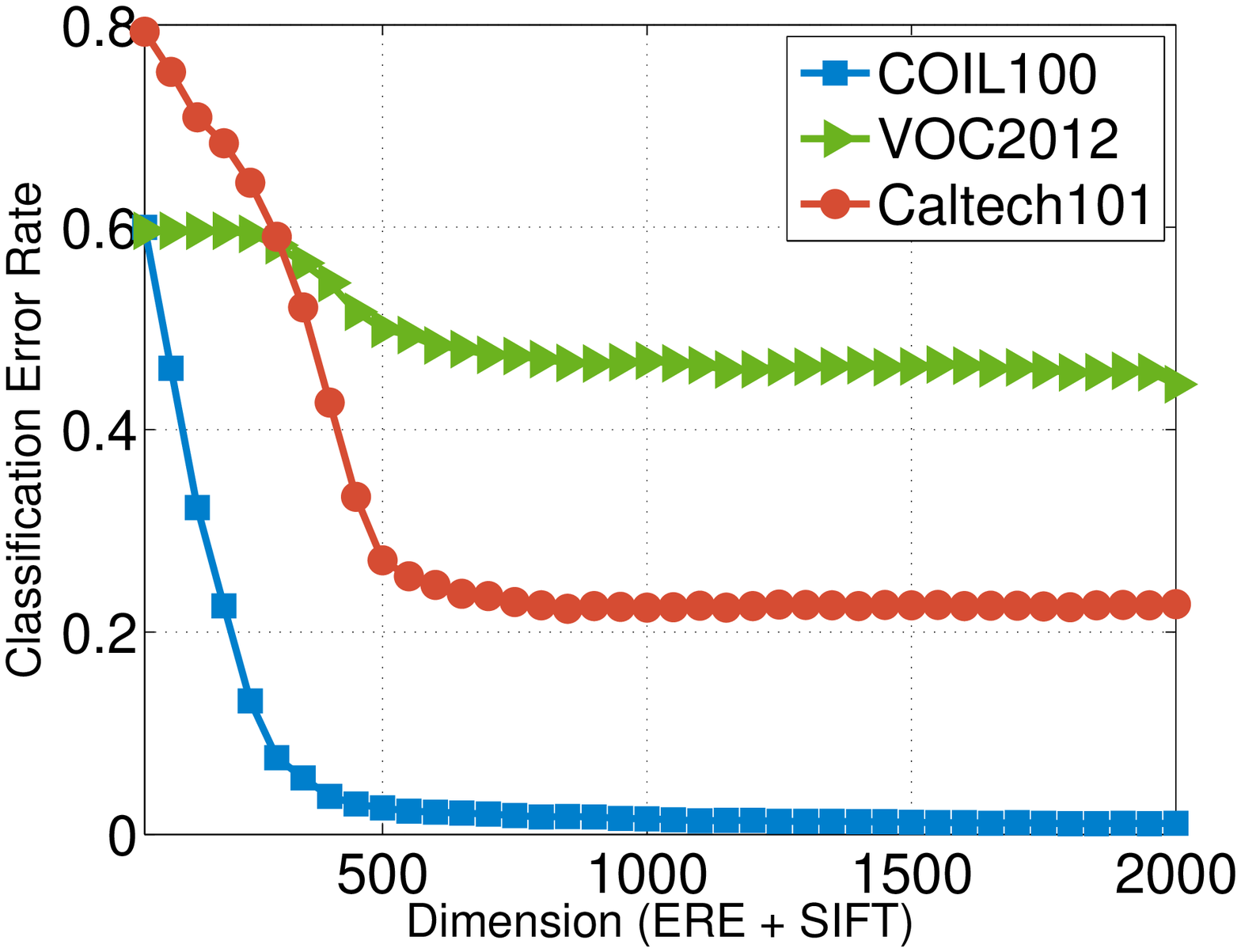}
\includegraphics[width=0.19\textwidth]{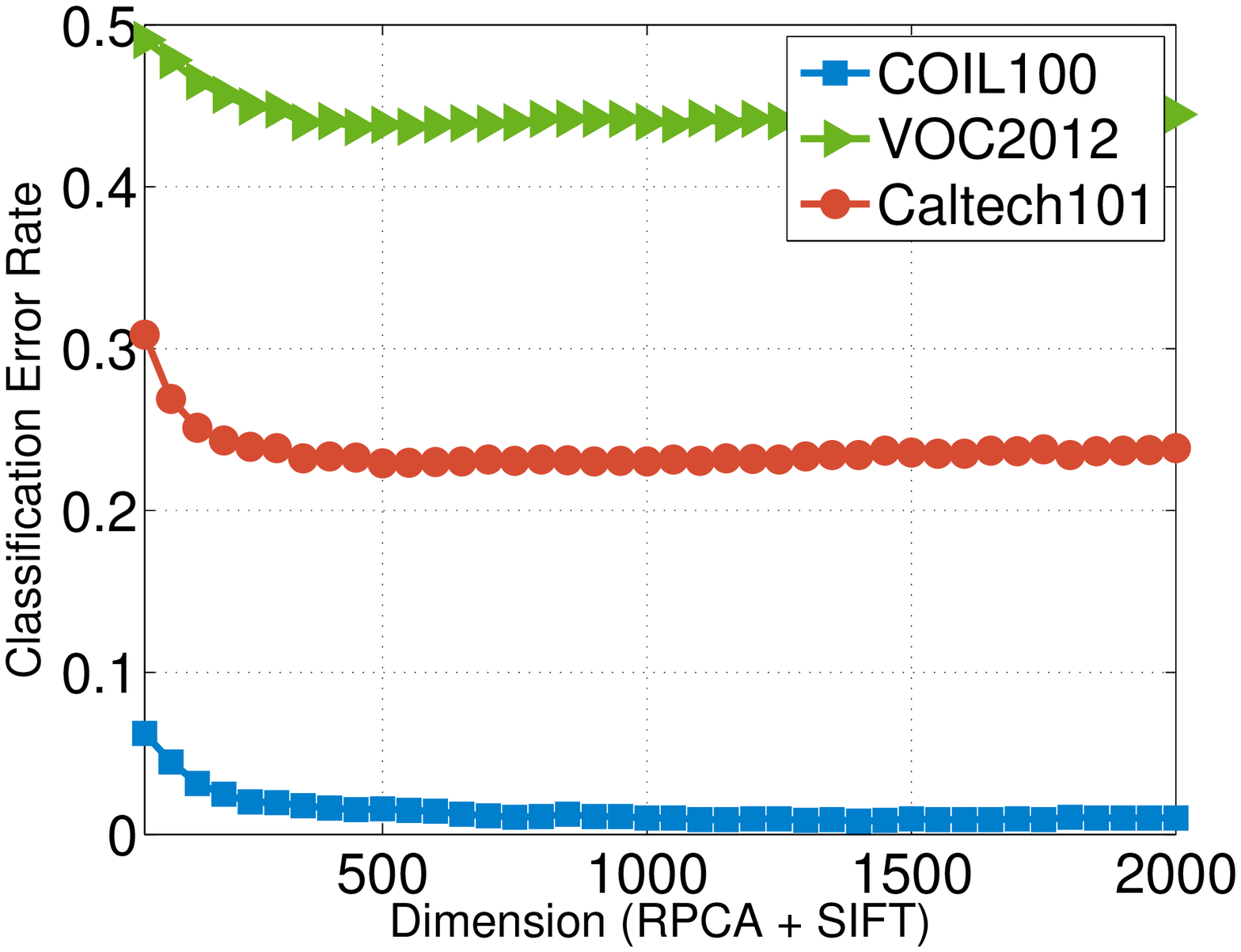}
\includegraphics[width=0.19\textwidth]{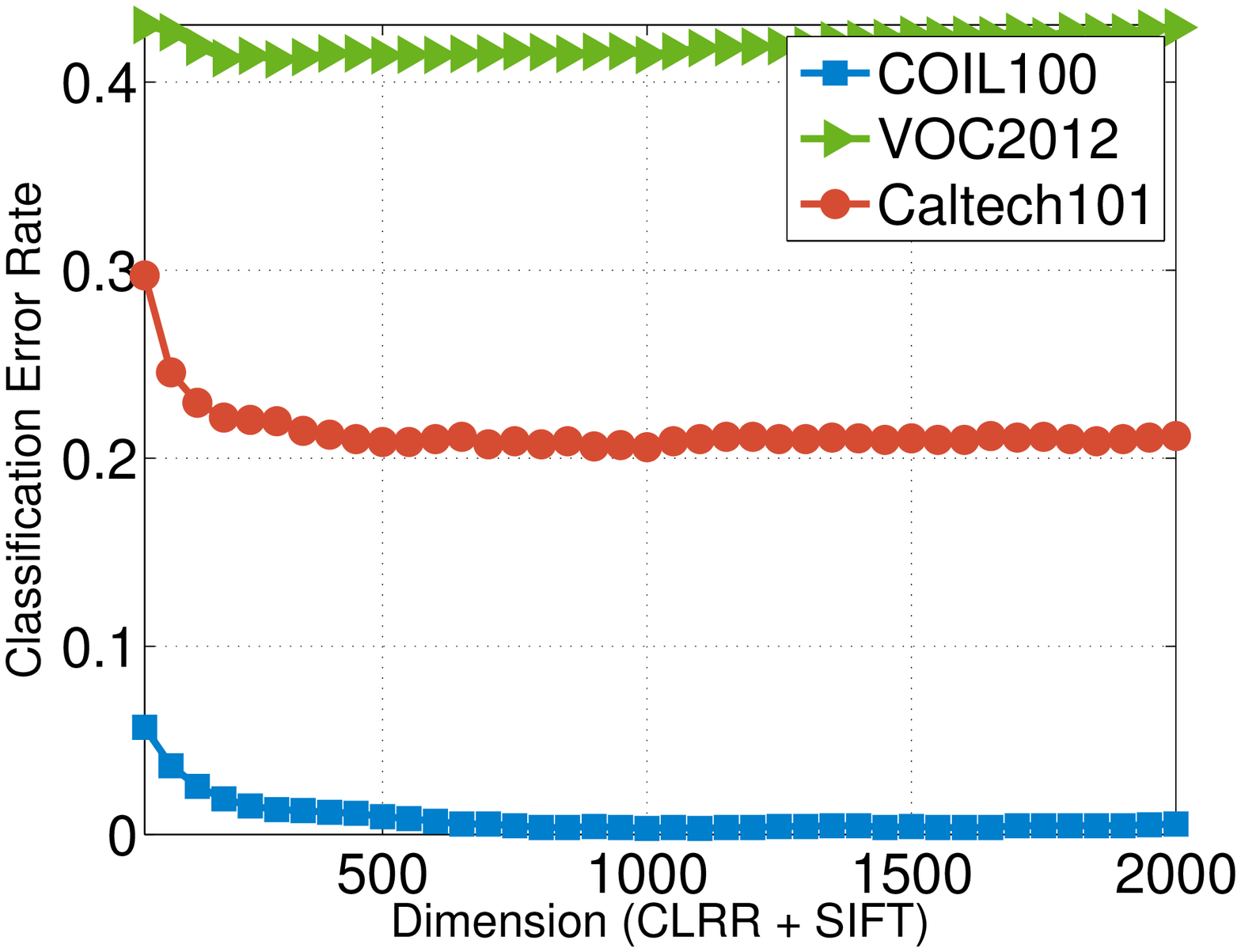}
\includegraphics[width=0.19\textwidth]{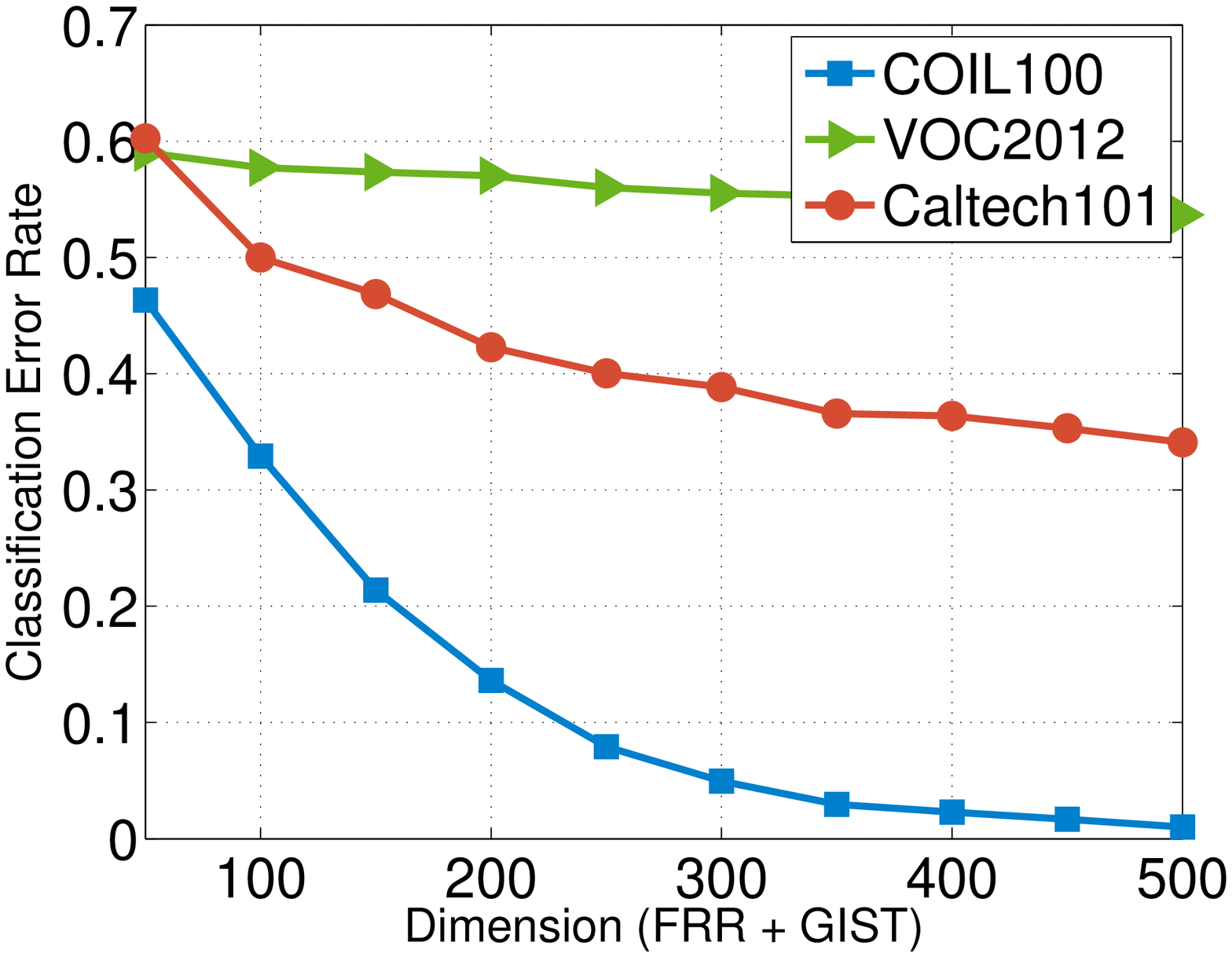}
\includegraphics[width=0.19\textwidth]{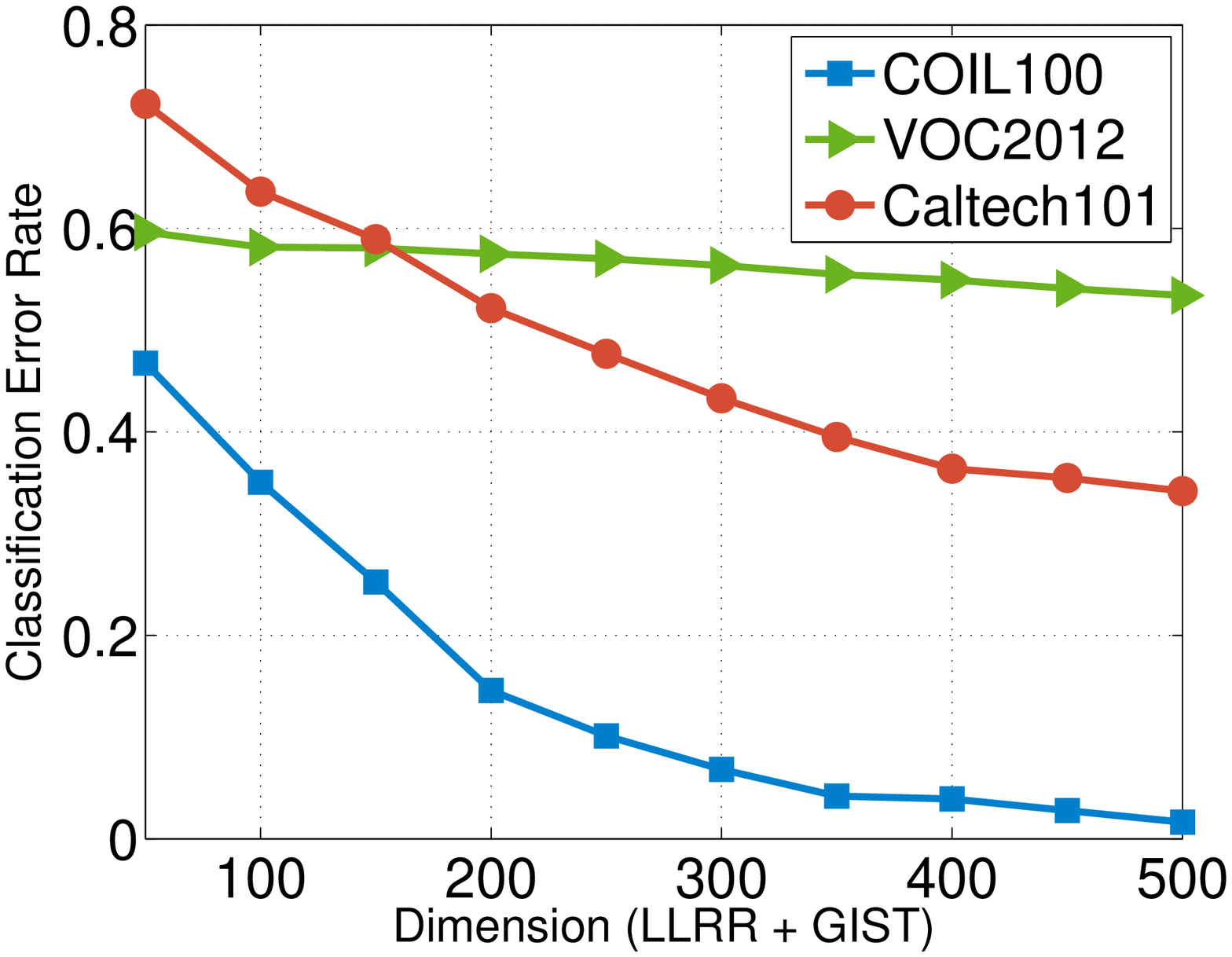}
\caption{Classification performance against different dimensions of three image databases. Others follow the same fashion.}
\label{fig-feaDimSel}
\end{figure*}

To have an overview of the parameter sensitivity on compared methods, we plot several parameter selection results in Figure~\ref{fig-methodParaSel}. This intuitively shows the influences of parameters on classification performance. From these figures, we observe that most parameters enjoy promising performances in a wide range on different databases. In practice, it is advised to choose the optimal parameters considering both the searching cost and the generalization ability of the method.

To show the convergence of iteration-based methods, we draw the convergence curves in Figure~\ref{fig-converge}. As vividly depicted in the figures, CLRR converges fast in a few iterations on different databases as others do. While CLRR and SRRS are at the same level of computational complexity higher than other low-rank approaches, they enjoy more inspiring performances on image classification, which could compensate for the larger computing overheads. In particular, CLRR not only performs satisfactorily on classification, but also can be used for regression while the rest would fail. It should be acknowledged that ERE and SPCA indeed run fast compared with others, however, there exists a shortcoming that they can neither be applied to regression nor to robust recovery on noisy data, which can limit their potential applications.

Furthermore, we illustrate several examples of choosing the optimal parameter $C$ of the $C$-SVC model for LibSVM in Figure~\ref{fig-svmParaSel}. As shown in the charts, LibSVM exhibits robustness in a wider range of parameter space on VOC2012 and Caltech101 than COIL100, and larger values are preferred as the proper $C$. Apart from that, we have provided several charts to show the classification performance using LibSVM against different dimensions of the data in Figure~\ref{fig-feaDimSel}. From these curves, it can be seen that the performance of the reduced feature (SIFT) would reach the optimal with less dimensions, while that of the original feature (GIST) needs more dimensions. In some sense, this suggests that the leading dimensions of the reduced feature would dominate the data representation, and thus more dimensions are unable to further boost the performance.

\begin{figure}[!t]
\centering
\includegraphics[width=0.23\textwidth]{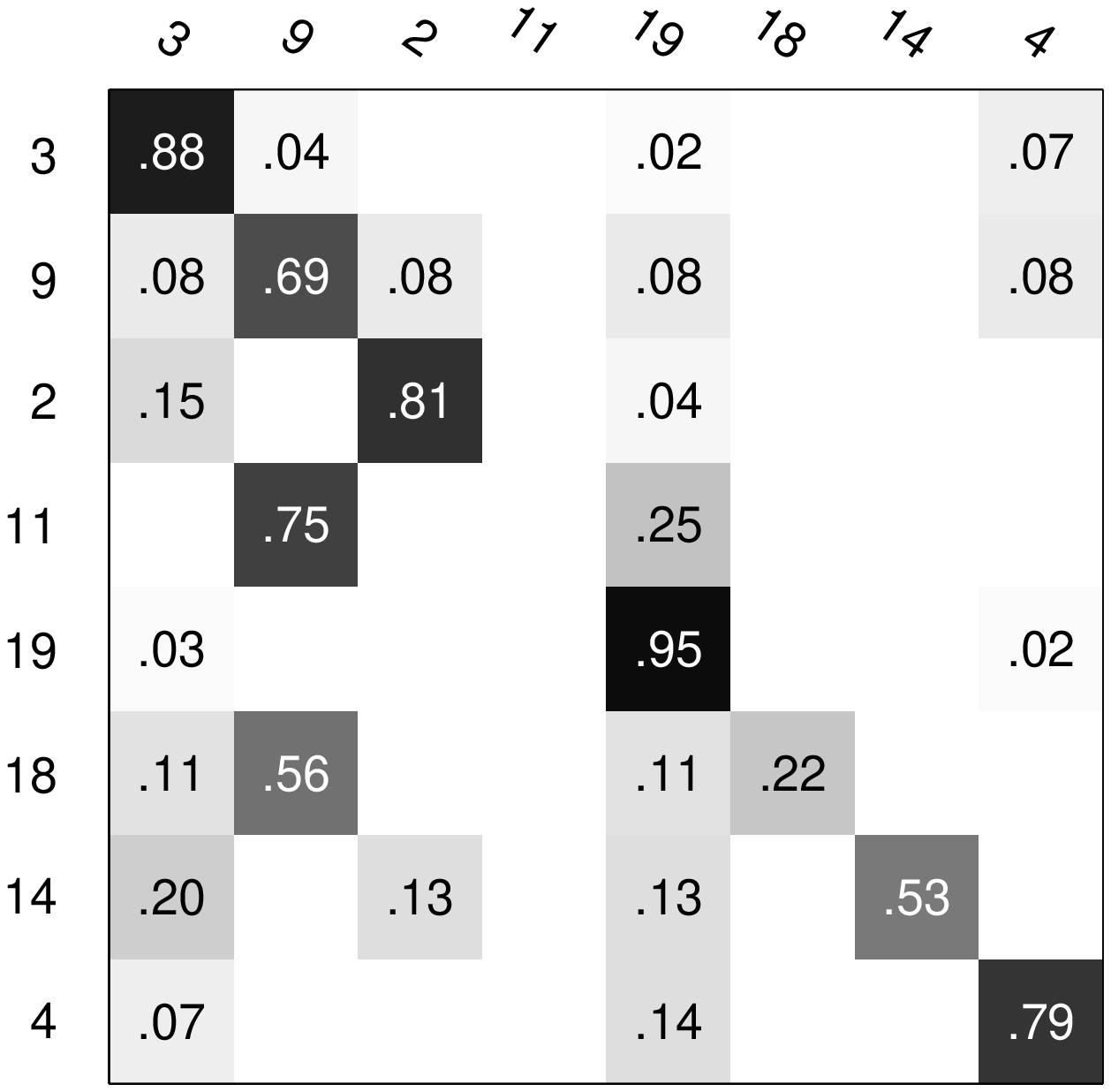} \hspace{0.1in}
\includegraphics[width=0.23\textwidth]{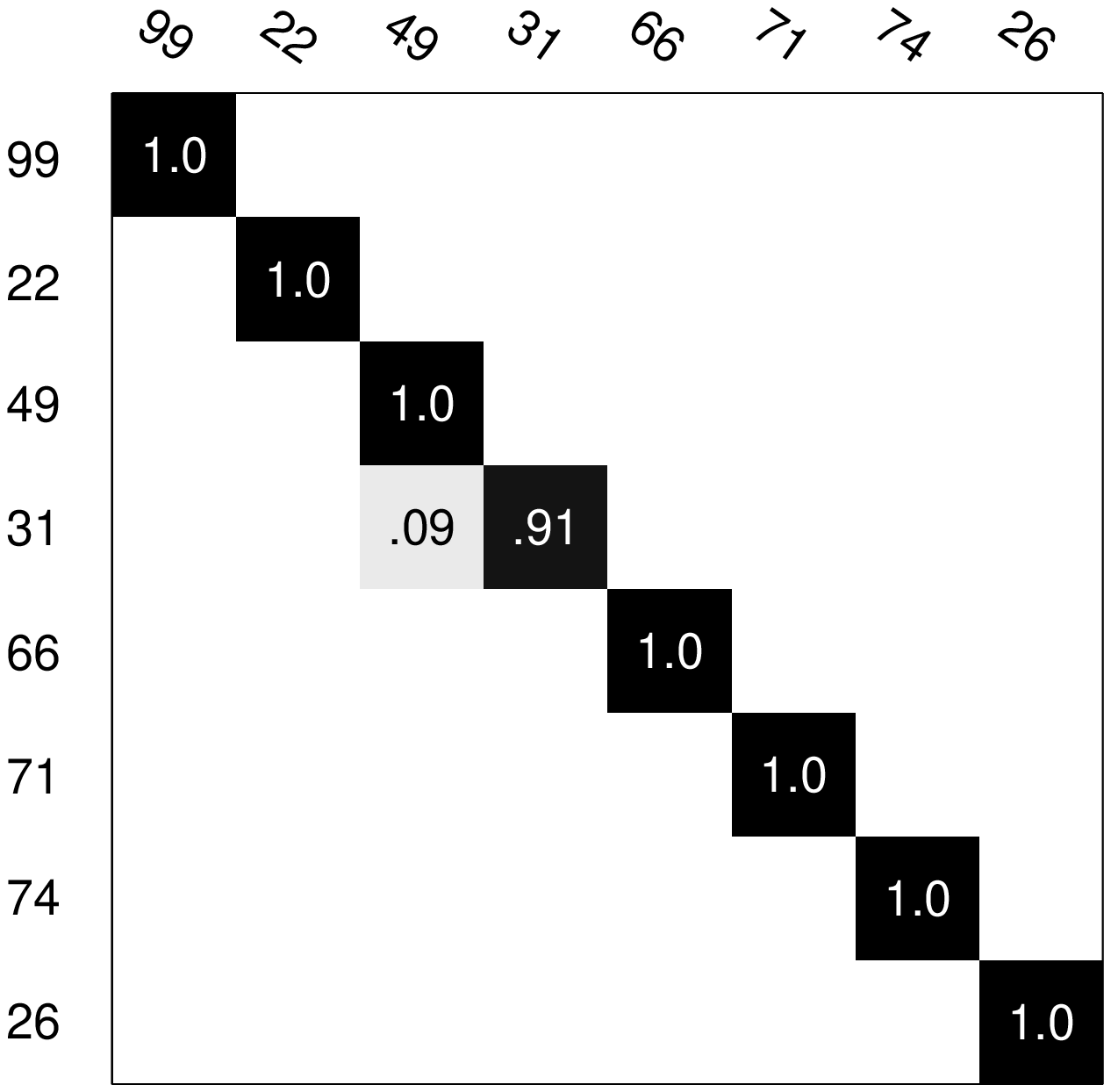}
\caption{Sample confusion matrices by performing CLRR on HOG feature of VOC2012 and Caltech101 using LibSVM. Eight categories were randomly chosen for display. The horizontal axis and the vertical axis denotes the paired categories in the database. The higher the diagonal values, the better the classification performance. It can be seen that it is very difficult to correctly classify the 11-$th$ class of VOC2012.}
\label{fig-CLRR-Confusion}
\vspace{-0.1in}
\end{figure}

In addition, we have displayed the confusion matrices generated by randomly choosing eight categories from the classification results on VOC2012 and Caltech101 for CLRR in Figure~\ref{fig-CLRR-Confusion}. From the figures, it is easy to calculate the True Positive Rate and the False Positive Rate for each category. This also gives us a possible direction to improve the integrated classification performance. Other methods have similar behaviors.

\subsection{Human Pose Estimation}
To investigate the regression performance of the proposed CLRR method, we consider the problem of estimating 3D configurations of complex articulated objects from monocular images for applications requiring 3D human body pose analysis. The database is detailed in \cite{agarwal-pami2006-recovering}. We choose the image silhouettes, as they are more reliably extractable. And the shape context distributions is used to give rise to 100D features for each sample. The 3D body pose is recovered as a 54D vector, including three joint angles for each of the 18 major body joints. Just like \cite{agarwal-pami2006-recovering}, we simply regress the original motion capture-based training format in the form of Euler angles. In total, we have used 1,691 training points coming from seven sequences and 418 testing points from one sequence, while the silhouette descriptors have 100 dimensions.

\begin{figure}[!t]
\centering
\includegraphics[width=0.42\textwidth]{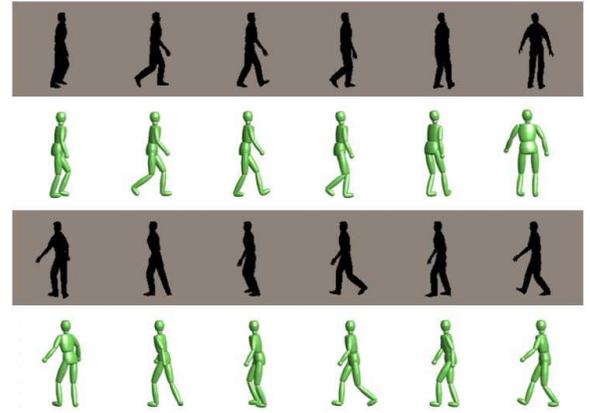}
\caption{Recovered 3D poses by CLRR and the corresponding silhouettes. The images and silhouettes have been normalized in scale for display purposes.}
\label{fig-pose}
\end{figure}

We compare CLRR with three popular regression methods including Least Squares Regression (LSR), Relevance Vector Machines (RVMs) \cite{tipping-jmlr2001-rvm} and Robust Regression (RR) \cite{Huang-eccv2012-robust}. The best parameters were obtained by five-fold cross-validation on training data. The results in terms of angle error are reported in Table~\ref{tbl-pose}. It can be observed that CLRR performs best, which is due to two reasons. On one hand, the robust projecting subspace derived from low-rank learning is more discriminative than LSR and RVM; on the other hand, CLRR respects the underlying data structure better than RR does, as RR implicitly assumes only a single subspace exists. In addition, we have drawn the recovered 3D body poses in Figure~\ref{fig-pose}, which shows a person walking from left to right and then in the inverse direction. We can see the recovered body poses are very close to the extracted silhouettes, demonstrating the favorable performance of our method.

%
\begin{table}[!t]
\centering
\caption{Angle error of human body estimation.}
\label{tbl-pose}
\begin{tabular}{|l|cccc|} \hline
Method   & LSR     & RVM    & RR   & CLRR     \\ \hline \hline
Angle error & 6.4931   & 6.0346  & 5.9168  & \bf{5.6927}  \\
\hline\end{tabular}
\end{table}

\subsection{Robust Face Recovery}
\begin{figure}[!t]
\centering
\includegraphics[width=0.48\textwidth]{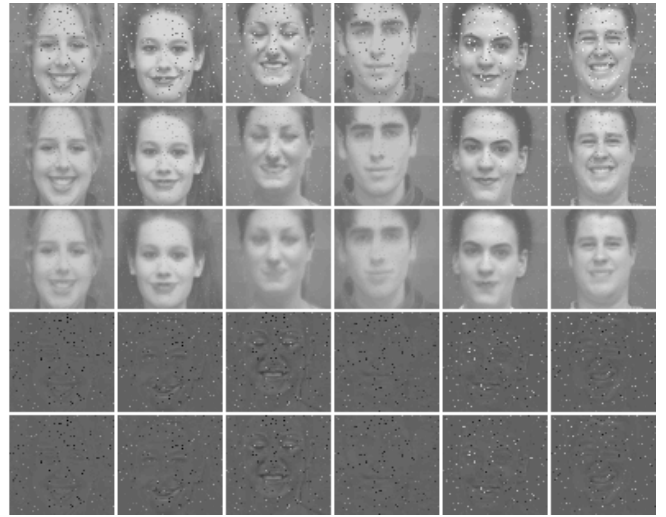}
\caption{Robust recovery performance on the CK+ database: row1- raw noisy image, row2- recovered image by LRR, row3 - recovered image by CLRR, row4 - error component $\mathbf{E}_i$ by LRR, row5 - error component $\mathbf{E}_i$ by CLRR.}
\label{fig-recovery}
\end{figure}

To examine the robustness of CLRR, we conducted experiments on the Cohn-Kanade AU-Coded Facial Expression Database\footnote{http://www.pitt.edu/~emotion/ck-spread.htm}\cite{kanade2000comprehensive,lucey-cvprw2010-CK+} In line with the user agreement, we randomly choose six available subjects including 848 images. They were cropped and resized to $64\times 64$. Here, we show the recovery effects of our method and LRR on several examples contaminated by artificial noise. To capture favorable performance, we have tuned the parameters to the best, and the resulting images are illustrated in Figure~\ref{fig-recovery}. These images indicate the superiority of our method to handle the noisy scenario. Specifically, the recovered faces by CLRR have much less noises compared with LRR. This is because the low-rank subspace $\mathbf{Z}$ is learned by simultaneously considering the Laplacian regularizer and the least squares regularizer, which encodes the supervised information in appropriate structures. Thus, the recovered faces from $\mathbf{XZ}$ are clearer compared with LRR. From another perspective, the noise of the faces is better encoded by the error matrix $\mathbf{E}$ when supervised information is employed in decomposing the noisy matrix into two components, \ie, the recovered component and the error component. Hence, these images have shown the effectiveness of the proposed method in the adverse situation where data points involve contaminations. In addition, the mouth of the fifth individual is closed by CLRR, which might be due to the reason that when learning the low-rank subspace, the samples would utilize the information of other samples within the same category. Furthermore, we observed that both LRR and CLRR share similar block-diagonal structures \wrt~six categories, which can be attributed to the fact that they are essentially low-rank learning approaches with only different constraints.
%
%

\section{Conclusions}
\label{sec7:conclusion}
This paper has developed a novel Constrained Low-Rank Representation method, which provides a sensible way to jointly learn both the discriminant lowest rank representation and the robust projecting subspace. Unlike most low-rank learning methods neglecting supervised information, we explicitly utilize label information via the adaptive least squares regularizer, such that the learned lowest rank representation and low-dimensional subspace have more discriminating power. Hence, it can naturally improve the performances of several real-world applications. The objective function is formulated as a constrained rank minimization problem solved by the inexact Augmented Lagrange Multiplier algorithm. Moreover, we have some discussions regarding the advantages, constraints and computational complexity analysis of CLRR. To demonstrate the effectiveness of our method, comprehensive experiments were conducted on image classification, human pose estimation, and robust face recovery. Results have clearly justified the promising efficacy of the proposed approach.

\ifCLASSOPTIONcompsoc
  \section*{Acknowledgments}
\else
  \section*{Acknowledgment}
\fi
The authors would like to thank the anonymous reviewers for their helpful and constructive comments that have greatly contributed to improving this manuscript.

\bibliographystyle{IEEEtran}
\bibliography{tcyb_clrr}

\ifCLASSOPTIONcaptionsoff
  \newpage
\fi

\end{document}